\documentclass{article}

\usepackage[final,nonatbib]{neurips_2019}

\usepackage[utf8]{inputenc} 
\usepackage[T1]{fontenc}    
\usepackage{hyperref}       
\usepackage{url}            
\usepackage{booktabs}       
\usepackage{amsfonts}       
\usepackage{nicefrac}       
\usepackage{microtype}      

\usepackage{graphicx}
\usepackage{subfigure}
\usepackage{enumitem}

\newcommand*\circled[1]{\tikz[baseline=(char.base)]{
            \node[shape=circle,draw,inner sep=.8pt] (char) {#1};}}

\usepackage[utf8]{inputenc}
\usepackage[T1]{fontenc}
\usepackage{textcomp}

\usepackage{tikz}
\usepackage{hyperref}
\usepackage{amsfonts}
\usepackage{amsmath}
\usepackage{float}
\usepackage{dblfloatfix}
\usepackage{cleveref}

\usepackage{multirow, makecell}
\usepackage{longtable}

\usepackage{etoolbox}
\usepackage{comment}
\usepackage{multirow}
\usepackage{xcolor}
\usepackage{algorithm}
\usepackage{algpseudocode}

\usepackage{caption}
\newcommand{\beginsupplement}{%
        \setcounter{table}{0}
        \renewcommand{\thetable}{S\arabic{table}}%
        \setcounter{figure}{0}
        \renewcommand{\thefigure}{S\arabic{figure}}%
     }

\newcommand{\ie}{\emph{i.e.}}
\newcommand{\eg}{\emph{e.g.}}

\newcommand{\yingyu}[1]{\textcolor{red}{Yingyu: #1}}

\newcommand{\todo}[1]{\textcolor{red}{#1}}

\usepackage{relsize}

\newtheorem{thm}{Theorem}
\newtheorem{definition}{Definition}

\title{N-Gram~Graph:~Simple~Unsupervised~Representation for Graphs, with Applications to Molecules}

\author{Shengchao Liu, Mehmet Furkan Demirel, Yingyu Liang\\
Department of Computer Sciences, University of Wisconsin-Madison, Madison, WI\\
\{shengchao, demirel, yliang\}@cs.wisc.edu
}

\begin{document}

\maketitle

\begin{abstract}
Machine learning techniques have recently been adopted in various applications in medicine, biology, chemistry, and material engineering. An important task is to predict the properties of molecules, which serves as the main subroutine in many downstream applications such as virtual screening and drug design. Despite the increasing interest, the key challenge is to construct proper representations of molecules for learning algorithms. This paper introduces the N-gram graph, a simple unsupervised representation for molecules. The method first embeds the vertices in the molecule graph. It then constructs a compact representation for the graph by assembling the vertex embeddings in short walks in the graph, which we show is equivalent to a simple graph neural network that needs no training. The representations can thus be efficiently computed and then used with supervised learning methods for prediction. Experiments on 60 tasks from 10 benchmark datasets demonstrate its advantages over both popular graph neural networks and traditional representation methods. This is complemented by theoretical analysis showing its strong representation and prediction power.
\end{abstract}

\section{Introduction} \label{s:introduction}

Increasingly, sophisticated machine learning methods have been used in non-traditional application domains like medicine, biology, chemistry, and material engineering~\cite{chen2018rise,camacho2018next,ching2018opportunities,butler2018machine}.
This paper focuses on a prototypical task of predicting properties of molecules. 
A motivating example is virtual screening for drug discovery. Traditional physical screening for drug discovery (\ie, selecting molecules based on properties tested via physical experiments) is typically accurate and valid, but also very costly and slow. In contrast, virtual screening (\ie, selecting molecules based on predicted properties via machine learning methods) can be done in minutes for predicting millions of molecules. Therefore, it can be a good filtering step before the physical experiments, to help accelerate the drug discovery process and significantly reduce resource requirements. The benefits gained then depend on the prediction performance of the learning algorithms. 

A key challenge is that raw data in these applications typically are not directly well-handled by existing learning algorithms and thus suitable representations need to be constructed carefully. 
Unlike image or text data where machine learning (in particular deep learning) has led to significant achievements, the most common raw inputs in molecule property prediction problems provide only highly abstract representations of the chemicals (\ie, graphs on atoms with atom attributes). 

To address the challenge, various representation methods have been proposed, mainly in two categories.
The first category is chemical fingerprints, the most widely used feature representations in aforementioned domains. The prototype is the Morgan fingerprints~~\cite{morgan1965generation} (see \Cref{fig:molecule_graph_representation} for an example).
The second category is graph neural networks (GNN)~\cite{pmlr-v70-gilmer17a,altae2017low,kearnes2016molecular,schutt2017quantum,gilmer2017neural,xu2018powerful}. They view the molecules as graphs with attributes, and build a computational network tailored to the graph structure that constructs a embedding vector for the input molecule and feed into a predictor (classifier or regression model). The network is trained end-to-end on labeled data, learning the embedding and the predictor at the same time. 

These different representation methods have their own advantages and disadvantages. 
The fingerprints are simple and efficient to calculate. They are also unsupervised and thus each molecule can be computed once and used by different machine learning methods for different tasks.
Graph neural networks in principle are more powerful: they can capture comprehensive information for molecules, including the skeleton structure, conformational information, and atom properties; they are trained end-to-end, potentially resulting in better representations for prediction. On the other hand, they need to be trained via supervised learning with sufficient labeled data, and for a new task the representation needs to retrained. Their training is also highly non-trivial and can be computationally expensive. So a natural question comes up: \emph{can we combine the benefits by designing a simple and efficient unsupervised representation method with great prediction performance?}

To achieve this, this paper introduces an unsupervised representation method called \textbf{N-gram graph}. It views the molecules as graphs and the atoms as vertices with attributes.
It first embeds the vertices by exploiting their special attribute structure. Then, it enumerates n-grams in the graph where an n-gram refers to a walk of length $n$, and constructs the embedding for each n-gram by assembling the embeddings of its vertices. The final representation is constructed based on the embeddings of all its n-grams. 
We show that the graph embedding step can also be formulated as a simple graph neural network that has no parameters and thus requires no training.
The approach is efficient, produces compact representations, and enjoys strong representation and prediction power shown by our theoretical analysis.
Experiments on 60 tasks from 10 benchmark datasets show that it gets overall better performance than both classic representation methods and several recent popular graph neural networks.

\paragraph{Related Work.}
We briefly describe the most related ones here due to space limitation and include a more complete review in~\Cref{s:related}.
Firstly, chemical fingerprints have long been used to represent molecules, including the classic Morgan fingerprints~\cite{morgan1965generation}.
They have recently been used with deep learning models~\cite{ma2015deep,unterthiner2014deep,liu2018practical,jastrzkebski2016learning,gomez2016automatic,kusner2017grammar}. 
Secondly, graph neural networks are recent deep learning models designed specifically for data with graph structure, such as social networks and knowledge graphs.
See~\Cref{s:background} for some brief introduction and refer to the surveys~\cite{hamilton2017representation,zhou2018graph,wu2019comprehensive} for more details. Since molecules can be viewed as structured graphs, various graph neural networks have been proposed for them. Popular ones include~\cite{altae2017low,kearnes2016molecular,schutt2017quantum,gilmer2017neural,xu2018powerful}. 
Finally, graph kernel methods can also be applied (\eg,~\cite{shawe2004kernel,shervashidze2011weisfeiler}). The implicit feature mapping induced by the kernel can be viewed as the representation for the input. The Weisfeiler-Lehman kernel~\cite{shervashidze2011weisfeiler} is particularly related due to its efficiency and theoretical backup. It is also similar in spirit to the Morgan fingerprints and closely related to the recent GIN graph neural network~\cite{xu2018powerful}.

\section{Preliminaries}  \label{s:short_pre}

\paragraph{Raw Molecule Data.} 
This work views a molecule as a graph, where each atom is a vertex and each bond is an edge.
Suppose there are $m$ vertices in the graph, denoted as $i \in \{0, 1, ..., m-1\}$.
Each vertex has useful attribute information, like the atom symbol and number of charges in the molecular graphs.
These vertex attributes are encoded into a vertex attribute matrix $\mathcal{V}$ of size $m \times S$, where $S$ is the number of attributes. An example of the attributes for vertex $i$ is:
\begin{align*}
    \mathcal{V}_{i,\cdot} = [\mathcal{V}_{i,0}, \mathcal{V}_{i,1}, \hdots, \mathcal{V}_{i,6}, \mathcal{V}_{i,7}] 
\end{align*}
where $\mathcal{V}_{i,0}$ is the atom symbol, $\mathcal{V}_{i,1}$ counts the atom degree, $\mathcal{V}_{i,6}$ and $\mathcal{V}_{i,7}$ indicate if it is an acceptor or a donor. Details are listed in \Cref{s:attribute_specification}.
Note that the attributes typically have discrete values. 
The bonding information is encoded into the adjacency matrix $\mathcal{A} \in \{0, 1\}^{m \times m}$, where  $\mathcal{A}_{i,j} = 1$ if and only if two vertices $i$ and $j$ are linked. We let $G = (\mathcal{V}, \mathcal{A})$ denote a molecular graph. 
Sometimes there are additional types of information, like bonding types and pairwise atom distance in the 3D Euclidean space used by~\cite{kearnes2016molecular,schutt2017quantum,gilmer2017neural}, which are beyond the scope of this work.

\paragraph{N-gram Approach.}
In natural language processing (NLP), an n-gram refers to a consecutive sequence of words. For example, the 2-grams of the sentence ``the dataset is large'' are $\{$``the dataset'', ``dataset is'', ``is large''$\}$. The N-gram approach constructs a representation vector $c_{(n)}$ for the sentence, whose coordinates correspond to all n-grams and the value of a coordinate is the number of times the corresponding n-gram shows up in the sentence. Therefore, the dimension of an n-gram vector is $|V|^n$ for a vocabulary $V$, and the vector $c_{(1)}$ is just the count vector of the words in the sentence. The n-gram representation has been shown to be a strong baseline (\eg,~\cite{wang2012baselines}). One drawback is its high dimensionality, which can be alleviated by using word embeddings. Let $W$ be a matrix whose $i$-th column is the embedding of the $i$-th word. Then $f_{(1)} = W c_{(1)}$ is just the sum of the word vectors in the sentence, which is in lower dimension and has also been shown to be a strong baseline (e.g.,~\cite{wieting2015towards,arora2016simple}).
In general, an n-gram can be embedded as the element-wise product of the word vectors in it.
Summing up all n-gram embeddings gives the embedding vector $f_{(n)}$.
This has been shown both theoretically and empirically to preserve good information for downstream learning tasks even using random word vectors (\eg,~\cite{arora2018acompressed}).

\section{N-gram Graph Representation}
\label{s:n_gram_method}

Our N-gram graph method consists of two steps: first embed the vertices, and then embed the graph based on the vertex embedding.

\subsection{Vertex Embedding} \label{sec:random_projection}

 \begin{figure*}[htb!]
    \centering
    \begin{tikzpicture}
\node[anchor=west, minimum height=20pt, minimum width=2cm] at (-2cm,2cm) {Neighbor};
\node[rectangle, fill={rgb:black,1;white,2}, anchor=west, minimum height=8pt, minimum width=0.1cm] at (0.2cm,2cm) {};
\node[rectangle, fill={rgb:black,1;white,2}, anchor=west, minimum height=8pt, minimum width=0.1cm] at (1.4cm,2cm) {};
\node[rectangle, fill={rgb:black,1;white,2}, anchor=west, minimum height=8pt, minimum width=0.1cm] at (3.4cm,2cm) {};
\node[rectangle, draw=black, anchor=west, minimum height=8pt, minimum width=1cm] at (0cm,2cm) (A1) {};
\node[rectangle, draw=black, anchor=west, minimum height=8pt, minimum width=1cm] at (1cm,2cm) (B1) {};
\node[rectangle, draw=black, anchor=west, minimum height=8pt, minimum width=1cm] at (2cm,2cm) (C1) {$\hdots$};
\node[rectangle, draw=black, anchor=west, minimum height=8pt, minimum width=1cm] at (3cm,2cm) (D1) {};

\node[anchor=west, minimum height=20pt, minimum width=2cm] at (-2cm,1.5cm) {Neighbor};
\node[rectangle, fill={rgb:black,1;white,2}, anchor=west, minimum height=8pt, minimum width=0.1cm] at (0cm,1.5cm) {};
\node[rectangle, fill={rgb:black,1;white,2}, anchor=west, minimum height=8pt, minimum width=0.1cm] at (1.6cm,1.5cm) {};
\node[rectangle, fill={rgb:black,1;white,2}, anchor=west, minimum height=8pt, minimum width=0.1cm] at (3.2cm,1.5cm) {};
\node[rectangle, draw=black, anchor=west, minimum height=8pt, minimum width=1cm] at (0cm,1.5cm) (A2) {};
\node[rectangle, draw=black, anchor=west, minimum height=8pt, minimum width=1cm] at (1cm,1.5cm) (B2) {};
\node[rectangle, draw=black, anchor=west, minimum height=8pt, minimum width=1cm] at (2cm,1.5cm) (C2) {$\hdots$};
\node[rectangle, draw=black, anchor=west, minimum height=8pt, minimum width=1cm] at (3cm,1.5cm) (D2) {};

\node[anchor=west, minimum height=20pt, minimum width=2cm] at (-2cm,0.5cm) {Neighbor};
\node[rectangle, fill={rgb:black,1;white,2}, anchor=west, minimum height=8pt, minimum width=0.1cm] at (0.4cm,0.5cm) {};
\node[rectangle, fill={rgb:black,1;white,2}, anchor=west, minimum height=8pt, minimum width=0.1cm] at (1.2cm,0.5cm) {};
\node[rectangle, fill={rgb:black,1;white,2}, anchor=west, minimum height=8pt, minimum width=0.1cm] at (3.6cm,0.5cm) {};
\node[rectangle, draw=black, anchor=west, minimum height=8pt, minimum width=1cm] at (0cm,0.5cm) (A3) {};
\node[rectangle, draw=black, anchor=west, minimum height=8pt, minimum width=1cm] at (1cm,0.5cm) (B3) {};
\node[rectangle, draw=black, anchor=west, minimum height=8pt, minimum width=1cm] at (2cm,0.5cm) (C3) {$\hdots$};
\node[rectangle, draw=black, anchor=west, minimum height=8pt, minimum width=1cm] at (3cm,0.5cm) (D3) {};

\draw[dotted] (C2) -- (C3) node[above][pos=0.8]{Padding};

\node[rectangle, draw=black, anchor=west, minimum height=8pt, minimum width=2cm] at (5cm,1.2cm) (embedded) {};
\draw[->] (D1.east) -- (embedded.north west) node[above][pos=0.3]{$W$};
\draw[->] (D2.east) -- (embedded.west) node[above][pos=0.3]{$W$};
\draw[->] (D3.east) -- (embedded.south west) node[above][pos=0.3]{$W$};
\node[anchor=west] at (5.5cm,1.6cm) {\protect\circled{1}};

\node[anchor=west, minimum height=20pt, minimum width=2cm] at (9cm,1.5cm) {Vertex $i$};
\node[rectangle, fill={rgb:black,1;white,2}, anchor=west, minimum height=8pt, minimum width=0.1cm] at (8.4cm,1.2cm) {};
\node[rectangle, fill={rgb:black,1;white,2}, anchor=west, minimum height=8pt, minimum width=0.1cm] at (9.6cm,1.2cm) {};
\node[rectangle, fill={rgb:black,1;white,2}, anchor=west, minimum height=8pt, minimum width=0.1cm] at (11cm,1.2cm) {};
\node[rectangle, draw=black, anchor=west, minimum height=8pt, minimum width=1cm] at (8cm,1.2cm) (A0) {};
\node[rectangle, draw=black, anchor=west, minimum height=8pt, minimum width=1cm] at (9cm,1.2cm) (B0) {};
\node[rectangle, draw=black, anchor=west, minimum height=8pt, minimum width=1cm] at (10cm,1.2cm) (C0) {$\hdots$};
\node[rectangle, draw=black, anchor=west, minimum height=8pt, minimum width=1cm] at (11cm,1.2cm) (D0) {};

\draw[->] (embedded.east) -- (A0.west) node[above][pos=0.5]{\protect\circled{2}};

    \end{tikzpicture}
    \scriptsize
    \caption{
        The CBoW-like neural network $g$.
        Each small box represents one attribute, and the gray color represents the bit one since it is one-hot encoded.
        Each long box consists of $S$ attributes with length $K$.
        \protect\circled{1} is the summation of all the embeddings of the neighbors of vertex $i$, where $W \in \mathbb{R}^{r \times K}$ is the vertex embedding matrix.
        \protect\circled{2} is a fully-connected neural network, and the final predictions are the attributes of vertex $i$.
s    }
    \label{fig:cbow_projection}
\end{figure*}
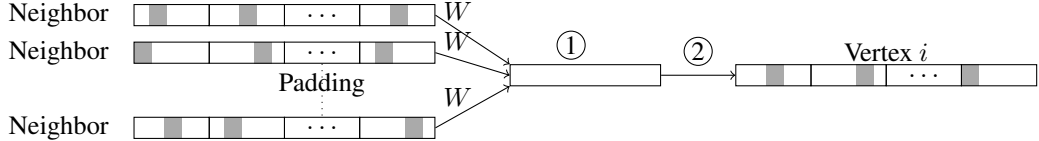

The typical method to embed vertices in graphs is to view each vertex as one token and apply an analog of CBoW~\cite{mikolov2013distributed} or other word embedding methods (\eg,~\cite{grover2016node2vec}).
Here we propose our variant that utilizes the structure that each vertex has several attributes of discrete values.\footnote{If there are numeric attributes, they can be simply padded to the learned embedding for the other attributes.} Recall that there are $S$ attributes; see~\Cref{s:short_pre}. Suppose the $j$-th attribute takes values in a set of size $k_j$, and let $K = \sum_{j=0}^{S-1} k_j$. Let $h^j_i$ denote a one-hot vector encoding the $j$-th attribute of vertex $i$, and let $h_i \in \mathbb{R}^K$ be the concatenation $h_i = [h^0_i; \hdots; h^{S-1}_i]$. Given an embedding dimension $r$, we would like to learn matrices $W^j \in \mathbb{R}^{r \times k_j}$ whose $\ell$-th column is an embedding vector for the $\ell$-th value of the $j$-th attribute. Once they are learned, we let $W \in \mathbb{R}^{r \times K}$ be the concatenation $W = [W^0, W^1, \hdots, W^{S-1}]$, and define the representation for vertex $i$ as
\begin{align} \label{eq:vertex_embedding_exp}
    f_i = W h_i. 
\end{align} 
Now it is sufficient to learn the vertex embedding matrix $W$. We use a CBoW-like pipeline; see Algorithm~\ref{alg:vertex_embedding}.  The intuition is to make sure the attributes $h_i$ of a vertex $i$ can be predicted from the $h_j$'s in its neighborhood. Let $C_i$ denote the set of vertices linked to $i$. We will train a neural network $\hat{h}_i = g(\{h_j: j\in C_i\})$ so that its output matches $h_i$.
As specified in Figure~\ref{fig:cbow_projection}, 
the network $g$ first computes $\sum_{j \in C_i} W h_j$ and then goes through a fully connected network with parameter $\theta$ to get $\hat{h}_i$. Given a dataset $\mathcal{S} = \{G_j =(\mathcal{V}_j, \mathcal{A}_j)\}$, the training is by minimizing the cross-entropy loss:
\begin{align}\label{eq:vertex_embedding_loss}
    \min_{W, \theta}
    \sum_{G \in \mathcal{S}} \sum_{i \in G} \sum_{0 \le \ell < S} \text{cross-entropy}(h_i^\ell, \hat{h}_i^\ell),
    \text{  subject to  }
    [\hat{h}_i^0; \hdots; \hat{h}_i^{S-1}] = g(\{h_j: j\in C_i\}).
\end{align}
Note that this requires no labels, i.e., it is unsupervised. In fact, $W$ learned from one dataset can be used for another dataset. Moreover, even using random vertex embeddings can give reasonable performance. See \Cref{sec:experiment} for more discussions.

\begin{minipage}[t]{0.5\textwidth}
\vspace{0pt}
\begin{algorithm}[H]
    \centering
    \footnotesize
    \caption{Vertex Embedding}\label{alg:vertex_embedding}
    \begin{flushleft}
        {\bfseries Input:} Graphs $\mathcal{S} = \{G_j =(\mathcal{V}_j, \mathcal{A}_j)\}$
    \end{flushleft}
    
    \begin{algorithmic}[1]
        \For{each graph $G$ in the dataset $\mathcal{S}$}
            \For{each vertex $i$ in graph $G$}
                \State{Extract neighborhood context $C_i$}
            \EndFor
        \EndFor
        \State{Train the network $g$ via \Cref{eq:vertex_embedding_loss}, using the extracted contexts}
    \end{algorithmic}
    \begin{flushleft}
        {\bfseries Output:} vertex embedding matrix $W$
    \end{flushleft}
    
\end{algorithm}
\end{minipage}
\hfill
\begin{minipage}[t]{0.45\textwidth}
\vspace{0pt}
\begin{algorithm}[H]
    \centering
    \footnotesize
    \caption{Graph Embedding}\label{alg:graph_embedding}
        \begin{flushleft}
        {\bfseries Input:} Graph $G =(\mathcal{V}, \mathcal{A})$; vertex embedding matrix $W$; step $T$
        \end{flushleft}
    \begin{algorithmic}[1]
        \State{Use \Cref{eq:vertex_embedding_exp} on $W$ and $\mathcal{V}$ to compute $f_i$'s }
        \State{$F_{(1)} = F = [f_1, \hdots, f_m], f_{(1)} = F_{(1)} \mathbf{1}$}
        \For{each $n \in [2, T]$}
            \State{$F_{(n)} = ( F_{(n-1)} \mathcal{A}) \odot F$}
            \State{$f_{(n)} = F_{(n)} \mathbf{1}$}
        \EndFor
    \end{algorithmic}
    \begin{flushleft}
        {\bfseries Output:} $f_G = [ f_{(1)}; \hdots; f_{(T)}]$
    \end{flushleft}
    
\end{algorithm}
\end{minipage}

\subsection{Graph Embedding} \label{sec:n_gram_graph}



The N-gram graph method is inspired by the N-gram approach in NLP, extending it from linear graphs (sentences) to general graphs (molecules). It views the graph as a Bag-of-Walks and builds representations on them. 
Let an n-gram refer to a walk of length $n$ in the graph, and the n-gram walk set refer to the set of all walks of length $n$. 
The embedding $f_p \in \mathbb{R}^r$ of an n-gram $p$ is simply the element-wise product of the vertex embeddings in that walk. The embedding $f_{(n)}  \in \mathbb{R}^r$ for the n-gram walk set is defined as the sum of the embeddings for all n-grams. The final N-gram graph representation up to length $T$ is denoted as $f_G  \in \mathbb{R}^{Tr}$, and defined as the concatenation of the embeddings of the n-gram walk sets for $n \in \{1, 2, \hdots, T\}$. Formally, given the vertex embedding $f_i$ for vertex $i$, 
\begin{align} \label{eq:n_gram_walk}
    f_p & = \prod_{i \in p} f_i,
    \quad
    f_{(n)} = \sum_{p: \text{n-gram}} f_p,
    \quad
    f_G = [ f_{(1)}; \hdots; f_{(T)}],
\end{align}
where $\prod$ is the Hadamard product (element-wise multiplication), i.e., if $p=(1, 2, 4)$, then $f_p = f_1 \odot f_2 \odot f_4$. 

Now we show that the above Bag-of-Walks view is equivalent to a simple graph neural network in~\Cref{alg:graph_embedding}. Each vertex will hold a latent vector. The latent vector for vertex $i$ is simply initialized to be its embedding $f_i$. 
At iteration $n$, each vertex updates its latent vector by element-wise multiplying it with the sum of the latent vectors of its neighbors. Therefore, at the end of iteration $n$, the latent vector on vertex $i$ is the sum of the embeddings of the walks that ends at $i$ and has length $n$, and the sum of the all latent vectors is the embedding of the n-gram walk set (with proper scaling).
Let $F_{(n)}$ be the matrix whose $i$-th column is the latent vector on vertex $i$ at the end of iteration $n$, then we have~\Cref{alg:graph_embedding} for computing the N-gram graph embeddings. Note that this simple GNN has no parameters and needs no training. The run time is $O(rT(m+m_e))$ where $r$ is the vertex embedding dimension, $T$ is the walk length, $m$ is the number of vertices, and $m_e$ is the number of edges.

By construction, N-gram graph is permutation invariant, i.e., invariant to permutations of the orders of atoms in the molecule. Also, it is unsupervised, so can be used for different tasks on the same dataset, and with different machine learning models. More properties are discussed in the next section.

\section{Theoretical Analysis} \label{sec:analysis}

Our analysis follows the framework in~\cite{arora2018acompressed}. It shows that under proper conditions, the N-gram graph embeddings can recover the count statistics of walks in the graph, so there is a classifier on the embeddings competitive to any classifier on the count statistics. Note that typically the count statistics can recover the graph. So this shows the strong representation and prediction power.
Our analysis makes one mild simplifying assumption:
\begin{itemize}[topsep=0pt,partopsep=0pt,itemsep=0pt,parsep=0pt]
    \item For computing the embeddings, we exclude walks that contain two vertices with exactly the same attributes. 
\end{itemize}
This significantly simplifies the analysis. Without it, it is still possible to do the analysis but needs a complicated bound on the difference introduced by such walks. Furthermore, we conducted experiments on embeddings excluding such walks which showed similar performance (see~\Cref{s:walk_path_comparison}). So analysis under the assumption is sufficient to provide insights for our method.%
\footnote{We don't present the version of our method excluding such walks due to its higher computational cost.}

The analysis takes the Bayesian view by assuming some  prior on the vertex embedding matrix $W$. This approach has been used for analyzing word embeddings and verified by empirical observations~\cite{arora2016latent,arora2018linear,yin2018dimensionality,kasiviswanathan2019restricted}. To get some intuition, consider the simple case when we only have $S=1$ attribute and consider the 1-gram embedding $f_{(1)}$. Recall that 
$f_{(1)} = \sum_{p: 1\text{-gram}} \prod_{i \in p} f_i = \sum_{i} f_i = W \sum_{i} h_i$. Define $c_{(1)} := \sum_{i} h_i$ which is the count vector of the occurrences of different types of 1-grams (i.e., vertices) in the graph, and we have $f_{(1)} = W c_{(1)}$. It is well known that there are various prior distributions over $W$ such that it has the Restricted Isometry Property (RIP), and if additionally $c_{(1)}$ is sparse, then $c_{(1)}$ can be efficiently recovered by various methods in the field of compressed sensing~\cite{foucart2017mathematical}. This means that $f_{(1)}$ preserves the information in $c_{(1)}$. 
The preservation then naturally leads to the prediction power~\cite{calderbank2009compressed,arora2018acompressed}.
Such an argument can be applied to the general case when $S>1$ and $f_{(n)}$ with $n>1$. We summarize the results below and present the details in Appendix~\ref{s:proof_analysis}.

\paragraph{Representation Power.}
Given a graph, let us define the bag-of-n-cooccurrences vector $c_{(n)}$ as follows (slightly generalizing~\cite{arora2018acompressed}). Recall that $S$ is the number of attributes, and $K=\sum_{j=0}^{S-1} k_j$ where $k_j$ is the number of possible values for the $j$-th attribute, and the value on the $i$-th vertex is denoted as $\mathcal{V}_{i,j}$.
\begin{definition}
Given a walk $p=(i_1, \dots, i_n)$ of length $n$, the vector $\mathbf{e}_p^{(j)} \in \mathbb{R}^{ { k_j \choose n} }$ is defined as the one-hot vector for the $j$-th attribute values $\{\mathcal{V}_{i_1, j}, \dots, \mathcal{V}_{i_n, j} \}$ along the walk.
The bag-of-n-cooccurrences vector $c_{(n)}$ is the concatenation of $c_{(n)}^{(0)}, \dots, c_{(n)}^{(S-1)}$, where $c_{(n)}^{(j)} = \sum_{p} \mathbf{e}_p^{(j)}$ with the sum over all paths $p$ of length $n$.
Furthermore, let the count statistics $c_{[T]}$ be the concatenation of $c_{(1)}, \dots, c_{(T)}$.
\end{definition}
So $c_{(n)}^{(j)}$ is the histogram of different values of the $j$-th attribute along the path, and $c_{(n)}$ is a concatenation over all the attributes.
It is in high dimension $\sum_{j=0}^{S-1} { k_j \choose n}$.
The following theorem then shows that $f_{(n)}$ is a compressed version and preserves the information of bag-of-n-cooccurrences. 

\begin{thm} \label{thm:rep}
If $r = \Omega(n s_n^3 \log K)$ where $s_{n}$ is the sparsity of $c_{(n)}$, then there is a prior distribution over $W$ so that $f_{(n)} = {T}_{(n)} c_{(n)}$ for a linear mapping $T_{(n)}$. If additionally $c_{(n)}$ is the sparsest vector satisfying $f_{(n)} = {T}_{(n)} c_{(n)}$, then with probability $1 - O(S \exp(-(r/S)^{1/3}))$, $c_{(n)}$ can be efficiently recovered from $f_{(n)}$. 
\end{thm}
The sparsity assumption of $c_{(n)}$ can be relaxed to be close to the sparsest vector (e.g., dense but only a few coordinates have large values), and then $c_{(n)}$ can be approximately recovered. This assumption is justified by the fact that there are a large number of possible types of n-gram while only a fraction of them are presented frequently in a graph. 
The prior distribution on $W$ can be from a wide family of distributions; see the proof in \Cref{s:proof_analysis}. This can also help explain that using random vertex embeddings in our method can also lead to good prediction performance; see~\Cref{sec:experiment}. In practice, the $W$ is learned and potentially captures better similarities among the vertices. 

The theorem means that $f_G$ preserves the information of the count statistics $c_{(n)} (1 \le n \le T)$. Note that typically, there are no two graphs having exactly the same count statistics, so the graph $G$ can be recovered from $f_G$. For example, consider a linear graph $b-c-d-a$, whose 2-grams are $(b,c), (c,d), (d,a)$. From the 2-grams it is easy to reconstruct the graph. In such cases, $f_G$ can be used to recover $G$, i.e., $f_G$ has full representation power of $G$. 

\paragraph{Prediction Power.}
Consider a prediction task and let $\ell_\mathcal{D}(g)$ denote the risk of a prediction function $g$ over the data distribution $\mathcal{D}$. 
\begin{thm}\label{thm:pre}
Let $g_c$ be a prediction function on the count statistics $c_{[T]}$.  
In the same setting as in Theorem~\ref{thm:rep}, with probability $1 - O(TS \exp(- (r/S)^{1/3}))$, there is a function $g_f$ on the N-gram graph embeddings $f_G$ with risk $\ell_\mathcal{D}(g_f) = \ell_\mathcal{D}(g_c)$. 
\end{thm}

So there always exists a predictor on our embeddings that has performance as good as any predictor on the count statistics. As mentioned, in typical cases, the graph $G$ can be recovered from the counts. Then there is always a predictor as good as the best predictor on the raw input $G$. Of course, one would like that not only $f_G$ has full information but also the information is easy to exploit. Below we provide the desired guarantee for the standard model of linear classifiers with $\ell_2$-regularization.

Consider the binary classification task with the logistic loss function $\ell(g, y)$ where $g$ is the prediction and $y$ is the true label. Let $\ell_\mathcal{D}(\theta) = \mathbb{E}_{\mathcal{D}} [ \ell(g_\theta, y)]$ 
denote the risk of a linear classifier $g_\theta$ with weight vector $\theta$ over the data distribution $\mathcal{D}$. Let $\theta^*$ denote the weight of the classifier over $c_{[n]}$ minimizing $\ell_\mathcal{D}$.
Suppose we have a dataset $\{(G_i, y_i)\}_{i=1}^M$ i.i.d.\ sampled from $\mathcal{D}$, and $\hat{\theta}$ is the weight over $f_G$ which is learned via $\ell_2$-regularization with regularization coefficient $\lambda$:
\begin{align}
\hat{\theta} = \arg\min_{\theta}  \frac{1}{M} \sum_{i=1}^M \ell(\langle \theta, f_{G_i} \rangle, y_i)  + \lambda \| \theta \|_2.
\end{align}
\begin{thm} \label{thm:pre_linear}
Assume that $f_G$ is scaled so that $\|f_G\|_2 \le 1$ for any graph from $\mathcal{D}$. There exists a prior distribution over $W$, such that with $r = \Omega(\frac{n s_{\max}^3}{\epsilon^2} \log K)$ for $s_{\max} = \max\{ s_n: 1\le n \le T\}$ and appropriate choice of regularization coefficient, with probability $1 - \delta - O(TS \exp(- (r/S)^{1/3}))$, the $\hat{\theta}$ minimizing the $\ell_2$-regularized logistic loss over the N-gram graph embeddings $f_{G_i}$'s satisfies 
\begin{align}
\ell_{\mathcal{D}}(\hat{\theta}) \le \ell_{\mathcal{D}}(\theta^*) + O\left( \|\theta^*\|_2 \sqrt{\epsilon + \frac{1}{M}\log\frac{1}{\delta} } \right).
\end{align} 
\end{thm}
Therefore, the linear classifier over the N-gram embeddings learned via the standard $\ell_2$-regularization have performance close to the best one on the count statistics. In practice, the label may depend nonlinearly on the count statistics or the embeddings, so one prefers more sophisticated models. Empirically, we can show that indeed the information in our embeddings can be efficiently exploited by classical methods like random forests and XGBoost. 

\section{Experiments} \label{sec:experiment}


Here we evaluate the N-gram graph method on 60 molecule property prediction tasks, comparing with three types of representations: Weisfeiler-Lehman Kernel, Morgan fingerprints, and several recent graph neural networks.
The results show that N-gram graph achieves better or comparable performance to the competitors.

\noindent\textbf{Methods.}\footnote{The code is available at \url{https://github.com/chao1224/n_gram_graph}. Baseline implementation follows \cite{Fey/Lenssen/2019,Ramsundar-et-al-2019}.}
\Cref{tab:model_and_representation} lists the feature representation and model combinations.
Weisfeiler-Lehman (WL) Kernel \cite{shervashidze2011weisfeiler}, Support Vector Machine (SVM), Morgan Fingerprints, Random Forest (RF), and XGBoost (XGB) \cite{chen2016xgboost} are chosen since they are the prototypical representation and learning methods in these domains.
Graph CNN (GCNN) \cite{altae2017low}, Weave Neural Network (Weave) \cite{kearnes2016molecular}, 
and Graph Isomorphism Network (GIN) \cite{xu2018powerful} are end-to-end graph neural networks, which are recently proposed deep learning models for handling molecular graphs.

\begin{table}[htb!]
    \footnotesize
    \centering
    \caption{
    Feature representation for each different machine learning model.
    Both Morgan fingerprints and N-gram graph are used with Random Forest (RF) and XGBoost (XGB).
    }
    \label{tab:model_and_representation}
    \begin{tabular}{| c | c |}
        \hline
        Feature Representation & Model \\
        \hline
        \hline
        Weisfeiler-Lehman Graph Kernel & SVM\\
        Morgan Fingerprints & RF, XGB\\
        Graph Neural Network & GCNN, Weave, GIN\\
        N-Gram Graph & RF, XGB\\
        \hline
    \end{tabular}
\end{table}

\noindent\textbf{Datasets.}
We test 6 regression and 4 classification datasets, each with multiple tasks.
Since our focus is to compare the representations of the graphs, no transfer learning or multi-task learning is considered.
In other words, we are comparing each task independently, which gives us 28 regression tasks and 32 classification tasks in total.
See \Cref{tab:vertex_attribute_matrix} for a detailed description of the attributes for the vertices in the molecular graphs from these datasets.
All datasets are split into five folds and with cross-validation results reported as follows.

\begin{itemize}
    \item Regression datasets: Delaney \cite{delaney_esol:_2004}, Malaria \cite{gamo_thousands_2010}, CEP \cite{hachmann_harvard_2011}, QM7 \cite{blum2009970}, QM8 \cite{ramakrishnan2015electronic}, QM9 \cite{ruddigkeit2012enumeration}.
    \item Classification datasets: Tox21 \cite{tox21}, ClinTox \cite{gayvert2016data,artemov2016integrated}, MUV \cite{rohrer2009maximum}, HIV \cite{hiv_2017}.
\end{itemize}

\noindent\textbf{Evaluation Metrics.}
Same evaluation metrics are utilized as in \cite{wu2018moleculenet}.
Note that as illustrated in \Cref{s:task_specification}, labels are highly skewed for each classification task, and thus ROC-AUC or PR-AUC is used to measure the prediction performance instead of accuracy.

\noindent\textbf{Hyperparameters.}
We tune the hyperparameter carefully for all representation and modeling methods. More details about hyperparameters are provided in Section~\Cref{s:hyperparameter}.
The following subsections display results with the N-gram parameter $T=6$ and the embedding dimension $r=100$.


\begin{table}[htb!]
    \centering
    \scriptsize
    \caption{
        Performance overview: (\# of tasks with top-1 performance, \# of tasks with top-3 performance) is listed for each model and each dataset.
        For cases with no top-3 performance on that dataset are left blank.
        Some models are not well tuned or too slow and are left in ``-''.
        }
    \label{tab:overall_performance}
    \begin{tabular}{|c|c|c|c|c|c|c|c|c|c|c|}
\hline
Dataset&\# Task&Eval Metric&\makecell{WL\\SVM}&\makecell{Morgan\\RF}&\makecell{Morgan\\XGB}&GCNN&Weave&GIN&\makecell{N-Gram\\RF}&\makecell{N-Gram\\XGB}
\\
\hline
\hline
Delaney&1&RMSE& & & & &1, 1&--&0, 1&0, 1 \\
Malaria&1&RMSE& &1, 1& & & &--&0, 1&0, 1 \\
CEP&1&RMSE& &1, 1& & & &--&0, 1&0, 1 \\
QM7&1&MAE& & & & &0, 1&--&0, 1&1, 1 \\
QM8&12&MAE& &1, 4&0, 1&7, 12&2, 6&--&0, 2&2, 11 \\
QM9&12&MAE&--& &0, 1&4, 7&1, 8&--&0, 8&7, 12 \\
Tox21&12&ROC-AUC&0, 2&0, 7& &0, 2&0, 1& &3, 12&9, 12 \\
clintox&2&ROC-AUC&0, 1& & &1, 2&0, 1& & &1, 2 \\
MUV&17&PR-AUC&4, 12&5, 11&5, 11& & &0, 7&2, 4&1, 6 \\
HIV&1&ROC-AUC& &1, 1& & & & &0, 1&0, 1 \\
\hline
Overall&60&&4, 15&9, 25&5, 13&12, 23&4, 18&0, 7&5, 31&\textbf{21, 48} \\
\hline
    \end{tabular}
\end{table}

\begin{figure}[htb!]
    \small
    \centering
    
    \begin{subfigure}[ROC-AUC of the best models on Tox21 (Morgan+RF, GCNN, N-gram+XGB). Larger is better.]
    {\includegraphics[width=0.95\linewidth]{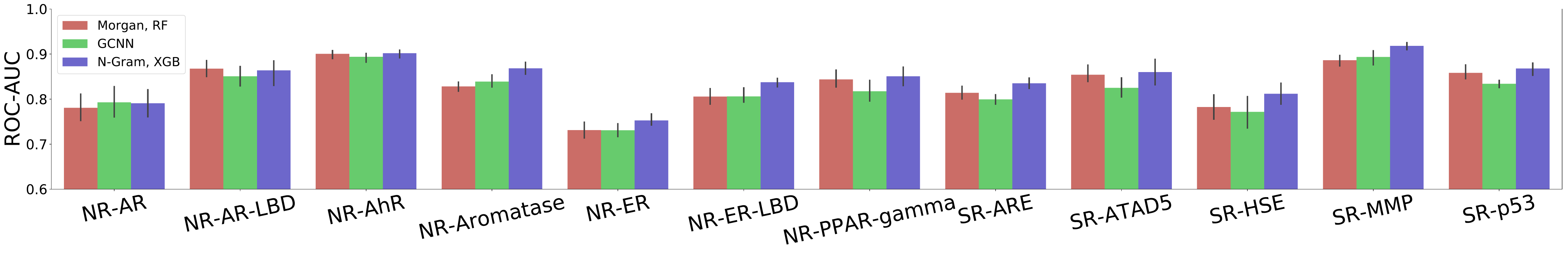}}
    \end{subfigure}\hfill

    \begin{subfigure}[MAE of the best models on QM9 (GCNN, Weave, N-gram+XGB). Smaller is better.]
    {\includegraphics[width=0.95\linewidth]{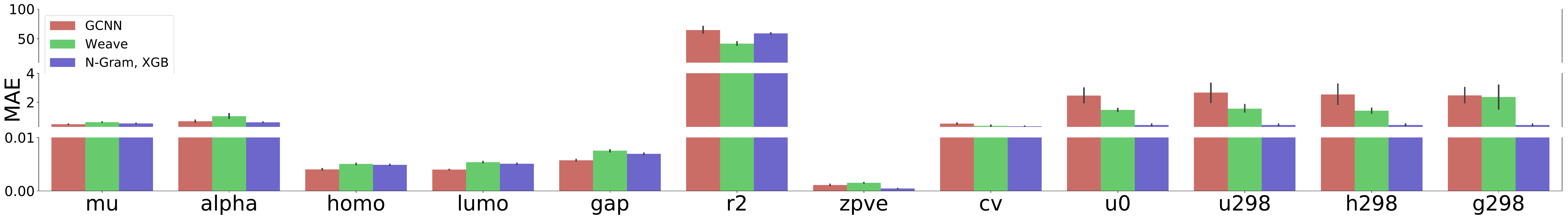}}
    \end{subfigure}\hfill

    \caption{
        Performance of the best models on the datasets Tox21 and QM9, averaged over 5-fold cross-validation.
    }
    \label{fig:performance}
\end{figure}

\paragraph{Performance.}
\Cref{tab:overall_performance} summarizes the prediction performance of the methods on all 60 tasks.
Since (1) no method can consistently beat all other methods on all tasks, and (2) for datasets like QM8, the error (MAE) of the best models are all close to 0, we report both the top-1 and top-3 number of tasks each method obtained. Such high-level overview can help better understand the model performance. Complete results are included in \Cref{s:complete_performance}.

Overall, we observe that N-gram graph, especially using XGBoost, shows better performance than the other methods. N-gram with XGBoost is in top-1 for 21 out of 60 tasks, and is in top-3 for 48. On some tasks, the margin is not large but the advantage is consistent; see for example the tasks on the dataset Tox21 in Figure~\ref{fig:performance}(a).
On some tasks, the advantage is significant; see for example the tasks u0, u298, h298, g298 on the dataset QM9 in Figure~\ref{fig:performance}(b).

We also observe that random forest on Morgan fingerprints has performance beyond general expectation, in particular, better than the recent graph neural network models on the classification tasks. 
One possible explanation is that we have used up to 4000 trees and obtained improved performance compared to 75 trees as in~\cite{wu2018moleculenet}, since the number of trees is the most important parameter as pointed out in \cite{liu2018practical}.
It also suggests that Morgan fingerprints indeed contains sufficient amount of information for the classification tasks, and methods like random forest are good at exploiting them.


\paragraph{Transferable Vertex Embedding.}
An intriguing property of the vertex embeddings is that they can be transferred across datasets. We evaluate N-Gram graph with XGB on Tox21, using  different vertex embeddings: trained on Tox21, random, and trained on other datasets.
See details in \Cref{s:transferable_vertex_embedding}.
\Cref{tab:transferrable} shows that embeddings from other datasets can be used to get comparable results. Even random embeddings can get good results, which is explained in~\Cref{sec:analysis}.
\begin{table}[htb!]
    \scriptsize
    \centering
    \caption{
        AUC-ROC of N-Gram graph with XGB on 12 tasks from Tox21.
        Six vertex embeddings are considered: non-transfer (trained on Tox21), vertex embeddings generated randomly and learned from 4 other datasets.
    }
    \label{tab:transferrable}
    \begin{tabular}{|c|c|c|c|c|c|c|}
\hline
&Non-Transfer&Random&Delaney&CEP&MUV&Clintox \\
\hline
NR-AR&0.791&0.790&0.785&0.787&0.796&0.780 \\
NR-AR-LBD&0.864&0.846&0.863&0.849&0.864&0.867 \\
NR-AhR&0.902&0.895&0.903&0.892&0.901&0.903 \\
NR-Aromatase&0.869&0.858&0.867&0.848&0.858&0.866 \\
NR-ER&0.753&0.751&0.752&0.740&0.735&0.747 \\
NR-ER-LBD&0.838&0.820&0.843&0.820&0.827&0.847 \\
NR-PPAR-gamma&0.851&0.809&0.862&0.813&0.832&0.857 \\
SR-ARE&0.835&0.823&0.841&0.814&0.835&0.842 \\
SR-ATAD5&0.860&0.830&0.844&0.817&0.845&0.857 \\
SR-HSE&0.812&0.777&0.806&0.768&0.805&0.810 \\
SR-MMP&0.918&0.909&0.918&0.902&0.916&0.919 \\
SR-p53&0.868&0.856&0.869&0.841&0.856&0.870 \\
\hline
    \end{tabular}
\end{table}

\paragraph{Computational Cost.}
\Cref{tab:representation_construction_time} depicts the representation construction time of different methods.
Since vertex embeddings can be amortized across different tasks on the same dataset or even transferred, the main runtime of our method is from the graph embedding step. It is relatively efficient, much faster than the GNNs and the kernel method, though Morgan fingerprints can be even faster.
\begin{table}[htb!]
\centering
\scriptsize
    \caption{
        Representation construction time in seconds.
        One task from each dataset as an example.
        Average over 5 folds, and including both the training set and test set.
    }
    \label{tab:representation_construction_time}
    \begin{tabular}{|c|c||c|c||c|c|c||c|c|}
\hline
Task&Dataset&\makecell{WL\\CPU}&\makecell{Morgan FPs\\CPU}&\makecell{GCNN\\GPU}&\makecell{Weave\\GPU}&\makecell{GIN\\GPU}&\makecell{Vertex, Emb\\GPU}&\makecell{Graph, Emb\\GPU}
\\
\hline
\hline
Delaney&Delaney&2.46&0.25&39.70&65.82&--&49.63&2.90 \\
Malaria&Malaria&128.81&5.28&377.24&536.99&--&1152.80&19.58 \\
CEP&CEP&1113.35&17.69&607.23&849.37&--&2695.57&37.40 \\
QM7&QM7&60.24&0.98&103.12&76.48&--&173.50&10.60 \\
E1-CC2&QM8&584.98&3.60&382.72&262.16&--&966.49&33.43 \\
mu&QM9&--&19.58&9051.37&1504.77&--&8279.03&169.72 \\
NR-AR&Tox21&70.35&2.03&130.15&142.59&608.57&525.24&10.81 \\
CT-TOX&Clintox&4.92&0.63&62.61&95.50&135.68&191.93&3.83 \\
MUV-466&MUV&276.42&6.31&401.02&690.15&1327.26&1221.25&25.50 \\
HIV&HIV&2284.74&17.16&1142.77&2138.10&3641.52&3975.76&139.85 \\
\hline
    \end{tabular}
\end{table}

\paragraph{Comparison to models using 3D information.}
What makes molecular graphs more complicated is that they contain 3D information, which is helpful for making predictions \cite{gilmer2017neural}.
Deep Tensor Neural Networks (DTNN) \cite{schutt2017quantum} and Message-Passing Neural Networks (MPNN) \cite{gilmer2017neural} are two graph neural networks that are able to utilize 3D information encoded in the datasets.%
\footnote{Weave~\cite{kearnes2016molecular} is also using the distance matrix, but it is the distance on graph, \ie the length of shortest path between each atom pair, not the 3D Euclidean distance.}
Therefore, we further compare our method to these two most advanced GNN models, on the two datasets QM8 and QM9 that have 3D information. The results are summarized in \Cref{tab:performance_3D}. 
The detailed results are in \Cref{tab:performance_regression_3d} and the computational times are in \Cref{tab:representation_construction_time_3d}.
They show that our method, though not using 3D information, still gets comparable performance.


\begin{table}[H]
    \centering
    \scriptsize
    \caption{
        Comparison of model using 3D information.
        On two regression datasets QM8 and QM9, and evaluated by MAE.
        N-Gram does not include any spatial information, like the distance between each atom pair, yet its performance is very comparative to the state-of-the-art methods.
    }
    \label{tab:performance_3D}
    \begin{tabular}{|c|c|c|c|c|c|c|c|c|c|c|}
\hline
Dataset&\# Task&\makecell{WL\\SVM}&\makecell{Morgan\\RF}&\makecell{Morgan\\XGB}&GCNN&Weave&DTNN&MPNN&\makecell{N-Gram\\RF}&\makecell{N-Gram\\XGB}
\\
\hline
\hline
QM8&12& &1, 4&0, 1&4, 10&0, 3&0, 5&5, 6&0, 2&2, 5 \\
QM9&12&--& & &0, 4&0, 1&7, 10&1, 9&0, 5&4, 7 \\
\hline
Overall&24& &1, 4&0, 1&4, 14&0, 4&7, 15&6, 15&0, 7&6, 12 \\
\hline
    \end{tabular}
\end{table}

\begin{figure}[H]
    \centering
    \includegraphics[width=1.\linewidth]{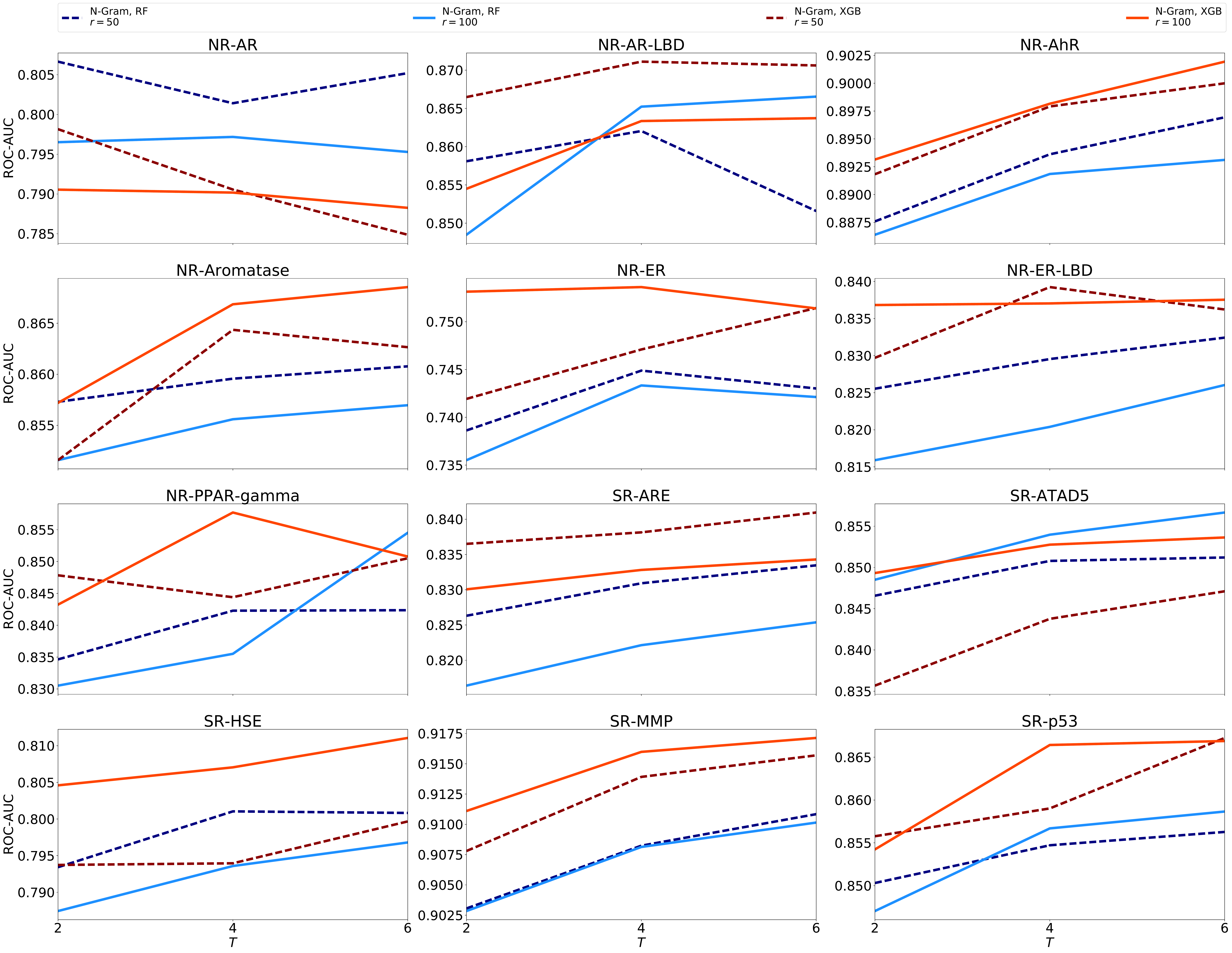}
    \caption{
        Effects of vertex embedding dimension $r$ and N-gram dimension $T$ on 12 tasks from Tox21: the effect of  $r$ and $T$ on ROC-AUC. $x$-axis: the hyperparameter $T$; $y$-axis: ROC-AUC. Different lines correspond to different methods and different values of $r$.
    }
    \label{fig:all_tox21}
\end{figure}

\paragraph{Effect of $r$ and $T$.}
We also explore the effect of the two key hyperparameters in N-Gram graph: the vertex embedding dimension $r$ and the N-gram length $T$. \Cref{fig:all_tox21} shows the results of 12 classification tasks on the Tox21 dataset are shown in, and \Cref{fig:all_rmse} shows the results on 3 regression tasks on the datasets Delaney, Malaria, and CEP. They reveal that generally, $r$ does not affect the model performance while increasing $T$ can bring in significant improvement.
More detailed discussions are in~\cref{s:effect_hyper}.

\section{Conclusion} \label{s:discuss}

This paper introduced a novel representation method called N-gram graph for molecule representation. It is simple, efficient, yet gives compact representations that can be applied with different learning methods. Experiments show that it can achieve overall better performance than prototypical traditional methods and several recent graph neural networks. 

The method was inspired by the recent word embedding methods and the traditional N-gram approach in natural language processing, and can be formulated as a simple graph neural network. It can also be used to handle general graph-structured data, such as social networks. 
Concrete future works include applications on other types of graph-structured data, pre-training and fine-tuning vertex embeddings, and designing even more powerful variants of the N-gram graph neural network.

\section*{Acknowledgements}
This work was supported in part by FA9550-18-1-0166. 
The authors would also like to acknowledge computing resources from the University of Wisconsin-Madison Center for High Throughput Computing and support provided by the University of Wisconsin-Madison Office of the Vice Chancellor for Research and Graduate Education with funding from the Wisconsin Alumni Research Foundation.

\bibliography{reference.bib}

\begin{thebibliography}{10}

\bibitem{hiv_2017}
Aids antiviral screen data.
\newblock
  \url{https://wiki.nci.nih.gov/display/NCIDTPdata/AIDS+Antiviral+Screen+Data}.
\newblock Accessed: 2017-09-27.

\bibitem{altae2017low}
Han Altae-Tran, Bharath Ramsundar, Aneesh~S Pappu, and Vijay Pande.
\newblock Low data drug discovery with one-shot learning.
\newblock {\em ACS Central Science}, 3(4):283--293, 2017.

\bibitem{arora2018acompressed}
Sanjeev Arora, Mikhail Khodak, Nikunj Saunshi, and Kiran Vodrahalli.
\newblock A compressed sensing view of unsupervised text embeddings,
  bag-of-n-grams, and lstm.
\newblock {\em International Conference on Learning Representations}, 2018.

\bibitem{arora2016latent}
Sanjeev Arora, Yuanzhi Li, Yingyu Liang, Tengyu Ma, and Andrej Risteski.
\newblock A latent variable model approach to pmi-based word embeddings.
\newblock {\em Transactions of the Association for Computational Linguistics},
  4:385--399, 2016.

\bibitem{arora2018linear}
Sanjeev Arora, Yuanzhi Li, Yingyu Liang, Tengyu Ma, and Andrej Risteski.
\newblock Linear algebraic structure of word senses, with applications to
  polysemy.
\newblock {\em Transactions of the Association of Computational Linguistics},
  6:483--495, 2018.

\bibitem{arora2016simple}
Sanjeev Arora, Yingyu Liang, and Tengyu Ma.
\newblock A simple but tough-to-beat baseline for sentence embeddings.
\newblock In {\em International Conference on Learning Representations}, 2016.

\bibitem{artemov2016integrated}
Artem~V Artemov, Evgeny Putin, Quentin Vanhaelen, Alexander Aliper, Ivan~V
  Ozerov, and Alex Zhavoronkov.
\newblock Integrated deep learned transcriptomic and structure-based predictor
  of clinical trials outcomes.
\newblock {\em bioRxiv}, page 095653, 2016.

\bibitem{blum2009970}
Lorenz~C Blum and Jean-Louis Reymond.
\newblock 970 million druglike small molecules for virtual screening in the
  chemical universe database gdb-13.
\newblock {\em Journal of the American Chemical Society}, 131(25):8732--8733,
  2009.

\bibitem{butler2018machine}
Keith~T Butler, Daniel~W Davies, Hugh Cartwright, Olexandr Isayev, and Aron
  Walsh.
\newblock Machine learning for molecular and materials science.
\newblock {\em Nature}, 559(7715):547, 2018.

\bibitem{calderbank2009compressed}
Robert Calderbank, Sina Jafarpour, and Robert Schapire.
\newblock Compressed learning: Universal sparse dimensionality reduction and
  learning in the measurement domain.
\newblock {\em Techical Report}, 2009.

\bibitem{camacho2018next}
Diogo~M Camacho, Katherine~M Collins, Rani~K Powers, James~C Costello, and
  James~J Collins.
\newblock Next-generation machine learning for biological networks.
\newblock {\em Cell}, 2018.

\bibitem{candes2008restricted}
Emmanuel~J Candes.
\newblock The restricted isometry property and its implications for compressed
  sensing.
\newblock {\em Comptes rendus mathematique}, 346(9-10):589--592, 2008.

\bibitem{candes2005decoding}
Emmanuel~J CANDES and Terence TAO.
\newblock Decoding by linear programming.
\newblock {\em IEEE transactions on information theory}, 51(12):4203--4215,
  2005.

\bibitem{chen2018rise}
Hongming Chen, Ola Engkvist, Yinhai Wang, Marcus Olivecrona, and Thomas
  Blaschke.
\newblock The rise of deep learning in drug discovery.
\newblock {\em Drug discovery today}, 2018.

\bibitem{chen2016xgboost}
Tianqi Chen and Carlos Guestrin.
\newblock Xgboost: A scalable tree boosting system.
\newblock In {\em Proceedings of the 22Nd ACM SIGKDD International Conference
  on Knowledge Discovery and Data Mining}, pages 785--794. ACM, 2016.

\bibitem{ching2018opportunities}
Travers Ching, Daniel~S Himmelstein, Brett~K Beaulieu-Jones, Alexandr~A
  Kalinin, Brian~T Do, Gregory~P Way, Enrico Ferrero, Paul-Michael Agapow,
  Michael Zietz, Michael~M Hoffman, et~al.
\newblock Opportunities and obstacles for deep learning in biology and
  medicine.
\newblock {\em Journal of The Royal Society Interface}, 15(141):20170387, 2018.

\bibitem{dahl2012deep}
George Dahl.
\newblock Deep learning how i did it: Merck 1st place interview.
\newblock {\em Online article available from http://blog. kaggle.
  com/2012/11/01/deep-learning-how-i-did-it-merck-1st-place-interview}, 2012.

\bibitem{delaney_esol:_2004}
John~S. Delaney.
\newblock {ESOL}: {Estimating} {Aqueous} {Solubility} {Directly} from
  {Molecular} {Structure}.
\newblock {\em Journal of Chemical Information and Computer Sciences},
  44(3):1000--1005, May 2004.

\bibitem{duvenaud_convolutional_nodate}
David~K Duvenaud, Dougal Maclaurin, Jorge Iparraguirre, Rafael Bombarell,
  Timothy Hirzel, Alan Aspuru-Guzik, and Ryan~P Adams.
\newblock Convolutional {Networks} on {Graphs} for {Learning} {Molecular}
  {Fingerprints}.
\newblock pages 2224--2232, 2015.

\bibitem{faber2017prediction}
Felix~A Faber, Luke Hutchison, Bing Huang, Justin Gilmer, Samuel~S Schoenholz,
  George~E Dahl, Oriol Vinyals, Steven Kearnes, Patrick~F Riley, and O~Anatole
  von Lilienfeld.
\newblock Prediction errors of molecular machine learning models lower than
  hybrid dft error.
\newblock {\em Journal of chemical theory and computation}, 13(11):5255--5264,
  2017.

\bibitem{Fey/Lenssen/2019}
Matthias Fey and Jan~E. Lenssen.
\newblock Fast graph representation learning with {PyTorch Geometric}.
\newblock In {\em ICLR Workshop on Representation Learning on Graphs and
  Manifolds}, 2019.

\bibitem{foucart2017mathematical}
Simon Foucart and Holger Rauhut.
\newblock A mathematical introduction to compressive sensing.
\newblock {\em Bull. Am. Math}, 54:151--165, 2017.

\bibitem{gamo_thousands_2010}
Francisco-Javier Gamo, Laura~M. Sanz, Jaume Vidal, Cristina~de Cozar, Emilio
  Alvarez, Jose-Luis Lavandera, Dana~E. Vanderwall, Darren V.~S. Green, Vinod
  Kumar, Samiul Hasan, James~R. Brown, Catherine~E. Peishoff, Lon~R. Cardon,
  and Jose~F. Garcia-Bustos.
\newblock Thousands of chemical starting points for antimalarial lead
  identification.
\newblock {\em Nature}, 465(7296):305--310, May 2010.

\bibitem{gayvert2016data}
Kaitlyn~M Gayvert, Neel~S Madhukar, and Olivier Elemento.
\newblock A data-driven approach to predicting successes and failures of
  clinical trials.
\newblock {\em Cell chemical biology}, 23(10):1294--1301, 2016.

\bibitem{pmlr-v70-gilmer17a}
Justin Gilmer, Samuel~S. Schoenholz, Patrick~F. Riley, Oriol Vinyals, and
  George~E. Dahl.
\newblock Neural message passing for quantum chemistry.
\newblock In Doina Precup and Yee~Whye Teh, editors, {\em Proceedings of the
  34th International Conference on Machine Learning}, volume~70 of {\em
  Proceedings of Machine Learning Research}, pages 1263--1272, International
  Convention Centre, Sydney, Australia, 06--11 Aug 2017. PMLR.

\bibitem{gilmer2017neural}
Justin Gilmer, Samuel~S Schoenholz, Patrick~F Riley, Oriol Vinyals, and
  George~E Dahl.
\newblock Neural message passing for quantum chemistry.
\newblock In {\em Proceedings of the 34th International Conference on Machine
  Learning-Volume 70}, pages 1263--1272. JMLR. org, 2017.

\bibitem{gomez2016automatic}
Rafael G{\'o}mez-Bombarelli, Jennifer~N Wei, David Duvenaud, Jos{\'e}~Miguel
  Hern{\'a}ndez-Lobato, Benjam{\'\i}n S{\'a}nchez-Lengeling, Dennis Sheberla,
  Jorge Aguilera-Iparraguirre, Timothy~D Hirzel, Ryan~P Adams, and Al{\'a}n
  Aspuru-Guzik.
\newblock Automatic chemical design using a data-driven continuous
  representation of molecules.
\newblock {\em ACS Central Science}, 2016.

\bibitem{grover2016node2vec}
Aditya Grover and Jure Leskovec.
\newblock node2vec: Scalable feature learning for networks.
\newblock In {\em Proceedings of the 22nd ACM SIGKDD international conference
  on Knowledge discovery and data mining}, pages 855--864. ACM, 2016.

\bibitem{hachmann_harvard_2011}
Johannes Hachmann, Roberto Olivares-Amaya, Sule Atahan-Evrenk, Carlos
  Amador-Bedolla, Roel~S. Sánchez-Carrera, Aryeh Gold-Parker, Leslie Vogt,
  Anna~M. Brockway, and Alán Aspuru-Guzik.
\newblock The {Harvard} {Clean} {Energy} {Project}: {Large}-{Scale}
  {Computational} {Screening} and {Design} of {Organic} {Photovoltaics} on the
  {World} {Community} {Grid}.
\newblock {\em The Journal of Physical Chemistry Letters}, 2(17):2241--2251,
  September 2011.

\bibitem{hamilton2017representation}
William~L Hamilton, Rex Ying, and Jure Leskovec.
\newblock Representation learning on graphs: Methods and applications.
\newblock {\em arXiv preprint arXiv:1709.05584}, 2017.

\bibitem{jastrzkebski2016learning}
Stanis{\l}aw Jastrz{\k{e}}bski, Damian Le{\'s}niak, and Wojciech~Marian
  Czarnecki.
\newblock Learning to smile (s).
\newblock {\em arXiv preprint arXiv:1602.06289}, 2016.

\bibitem{kasiviswanathan2019restricted}
Shiva~Prasad Kasiviswanathan and Mark Rudelson.
\newblock Restricted isometry property under high correlations.
\newblock {\em arXiv preprint arXiv:1904.05510}, 2019.

\bibitem{kearnes2016molecular}
Steven Kearnes, Kevin McCloskey, Marc Berndl, Vijay Pande, and Patrick Riley.
\newblock Molecular graph convolutions: moving beyond fingerprints.
\newblock {\em Journal of computer-aided molecular design}, 30(8):595--608,
  2016.

\bibitem{kusner2017grammar}
Matt~J Kusner, Brooks Paige, and Jos{\'e}~Miguel Hern{\'a}ndez-Lobato.
\newblock Grammar variational autoencoder.
\newblock {\em arXiv preprint arXiv:1703.01925}, 2017.

\bibitem{Landrum2016RDKit2016_09_4}
Greg Landrum.
\newblock Rdkit: Open-source cheminformatics software.
\newblock 2016.

\bibitem{li2015gated}
Yujia Li, Daniel Tarlow, Marc Brockschmidt, and Richard Zemel.
\newblock Gated graph sequence neural networks.
\newblock {\em arXiv preprint arXiv:1511.05493}, 2015.

\bibitem{liu2018practical}
Shengchao Liu, Moayad Alnammi, Spencer~S Ericksen, Andrew~F Voter, James~L
  Keck, F~Michael Hoffmann, Scott~A Wildman, and Anthony Gitter.
\newblock Practical model selection for prospective virtual screening.
\newblock {\em bioRxiv}, page 337956, 2018.

\bibitem{ma2015deep}
Junshui Ma, Robert~P Sheridan, Andy Liaw, George~E Dahl, and Vladimir Svetnik.
\newblock Deep neural nets as a method for quantitative structure--activity
  relationships.
\newblock {\em Journal of chemical information and modeling}, 55(2):263--274,
  2015.

\bibitem{matlock2018learning}
Matthew~K. Matlock, Na~Le Dang, and S.~Joshua Swamidass.
\newblock Learning a {Local}-{Variable} {Model} of {Aromatic} and {Conjugated}
  {Systems}.
\newblock {\em ACS Central Science}, 4(1):52--62, January 2018.

\bibitem{kaggle2012merck}
{Merck}.
\newblock Merck molecular activity challenge.
\newblock {\em https://www.kaggle.com/c/MerckActivity}, 2012.

\bibitem{mikolov2013distributed}
Tomas Mikolov, Ilya Sutskever, Kai Chen, Greg~S Corrado, and Jeff Dean.
\newblock Distributed representations of words and phrases and their
  compositionality.
\newblock In {\em Advances in neural information processing systems}, pages
  3111--3119, 2013.

\bibitem{morgan1965generation}
HL~Morgan.
\newblock The generation of a unique machine description for chemical
  structures-a technique developed at chemical abstracts service.
\newblock {\em Journal of Chemical Documentation}, 5(2):107--113, 1965.

\bibitem{ramakrishnan2015electronic}
Raghunathan Ramakrishnan, Mia Hartmann, Enrico Tapavicza, and O~Anatole
  Von~Lilienfeld.
\newblock Electronic spectra from tddft and machine learning in chemical space.
\newblock {\em The Journal of chemical physics}, 143(8):084111, 2015.

\bibitem{Ramsundar-et-al-2019}
Bharath Ramsundar, Peter Eastman, Patrick Walters, Vijay Pande, Karl Leswing,
  and Zhenqin Wu.
\newblock {\em Deep Learning for the Life Sciences}.
\newblock O'Reilly Media, 2019.
\newblock
  \url{https://www.amazon.com/Deep-Learning-Life-Sciences-Microscopy/dp/1492039837}.

\bibitem{rohrer2009maximum}
Sebastian~G Rohrer and Knut Baumann.
\newblock Maximum unbiased validation (muv) data sets for virtual screening
  based on pubchem bioactivity data.
\newblock {\em Journal of chemical information and modeling}, 49(2):169--184,
  2009.

\bibitem{ruddigkeit2012enumeration}
Lars Ruddigkeit, Ruud Van~Deursen, Lorenz~C Blum, and Jean-Louis Reymond.
\newblock Enumeration of 166 billion organic small molecules in the chemical
  universe database gdb-17.
\newblock {\em Journal of chemical information and modeling},
  52(11):2864--2875, 2012.

\bibitem{schutt2017quantum}
Kristof~T Sch{\"u}tt, Farhad Arbabzadah, Stefan Chmiela, Klaus~R M{\"u}ller,
  and Alexandre Tkatchenko.
\newblock Quantum-chemical insights from deep tensor neural networks.
\newblock {\em Nature communications}, 8:13890, 2017.

\bibitem{shawe2004kernel}
John Shawe-Taylor, Nello Cristianini, et~al.
\newblock {\em Kernel methods for pattern analysis}.
\newblock Cambridge university press, 2004.

\bibitem{shervashidze2011weisfeiler}
Nino Shervashidze, Pascal Schweitzer, Erik Jan~van Leeuwen, Kurt Mehlhorn, and
  Karsten~M Borgwardt.
\newblock Weisfeiler-lehman graph kernels.
\newblock {\em Journal of Machine Learning Research}, 12(Sep):2539--2561, 2011.

\bibitem{todeschini2009molecular}
Roberto Todeschini and Viviana Consonni.
\newblock {\em Molecular descriptors for chemoinformatics: volume I:
  alphabetical listing/volume II: appendices, references}, volume~41.
\newblock John Wiley \& Sons, 2009.

\bibitem{tox21}
{Tox21 Data Challenge}.
\newblock Tox21 data challenge 2014.
\newblock {\em https://tripod.nih.gov/tox21/challenge/}, 2014.

\bibitem{unterthiner2014deep}
Thomas Unterthiner, Andreas Mayr, G{\"u}nter Klambauer, Marvin Steijaert,
  J{\"o}rg~K Wegner, Hugo Ceulemans, and Sepp Hochreiter.
\newblock Deep learning as an opportunity in virtual screening.
\newblock {\em Advances in neural information processing systems}, 27, 2014.

\bibitem{wang2012baselines}
Sida Wang and Christopher~D Manning.
\newblock Baselines and bigrams: Simple, good sentiment and topic
  classification.
\newblock In {\em Proceedings of the 50th Annual Meeting of the Association for
  Computational Linguistics: Short Papers-Volume 2}, pages 90--94. Association
  for Computational Linguistics, 2012.

\bibitem{weininger1989smiles}
David Weininger, Arthur Weininger, and Joseph~L Weininger.
\newblock Smiles. 2. algorithm for generation of unique smiles notation.
\newblock {\em Journal of Chemical Information and Computer Sciences},
  29(2):97--101, 1989.

\bibitem{wieting2015towards}
John Wieting, Mohit Bansal, Kevin Gimpel, and Karen Livescu.
\newblock Towards universal paraphrastic sentence embeddings.
\newblock {\em arXiv preprint arXiv:1511.08198}, 2015.

\bibitem{wu2018moleculenet}
Zhenqin Wu, Bharath Ramsundar, Evan~N Feinberg, Joseph Gomes, Caleb Geniesse,
  Aneesh~S Pappu, Karl Leswing, and Vijay Pande.
\newblock Moleculenet: a benchmark for molecular machine learning.
\newblock {\em Chemical Science}, 9(2):513--530, 2018.

\bibitem{wu2019comprehensive}
Zonghan Wu, Shirui Pan, Fengwen Chen, Guodong Long, Chengqi Zhang, and Philip~S
  Yu.
\newblock A comprehensive survey on graph neural networks.
\newblock {\em arXiv preprint arXiv:1901.00596}, 2019.

\bibitem{xu2018powerful}
Keyulu Xu, Weihua Hu, Jure Leskovec, and Stefanie Jegelka.
\newblock How powerful are graph neural networks?
\newblock {\em arXiv preprint arXiv:1810.00826}, 2018.

\bibitem{yin2018dimensionality}
Zi~Yin and Yuanyuan Shen.
\newblock On the dimensionality of word embedding.
\newblock In {\em Advances in Neural Information Processing Systems}, pages
  895--906, 2018.

\bibitem{ying2018hierarchical}
Rex Ying, Jiaxuan You, Christopher Morris, Xiang Ren, William~L Hamilton, and
  Jure Leskovec.
\newblock Hierarchical graph representation learning withdifferentiable
  pooling.
\newblock {\em arXiv preprint arXiv:1806.08804}, 2018.

\bibitem{zhou2018graph}
Jie Zhou, Ganqu Cui, Zhengyan Zhang, Cheng Yang, Zhiyuan Liu, and Maosong Sun.
\newblock Graph neural networks: A review of methods and applications.
\newblock {\em arXiv preprint arXiv:1812.08434}, 2018.

\end{thebibliography}
\bibliographystyle{plain}

\clearpage

\beginsupplement
\appendix

\LTcapwidth=\textwidth

\newcolumntype{C}[1]{>{\centering\let\newline\\\arraybackslash\hspace{0pt}}m{#1}}

\section{Related Work} \label{s:related}

There are a large number of works along the line of machine learning for molecules and we review the more related ones here.

The adoption of sophisticated machine learning methods, in particular deep learning methods, has been recent trend in the domains of medicine, biology, chemistry, etc~\cite{chen2018rise,camacho2018next,ching2018opportunities,butler2018machine}.
Deep learning methods started to capture the attention among scientists in the drug discovery domain from Merck Molecular Activity Challange~\cite{kaggle2012merck,dahl2012deep}.
Efforts expanded to investigate the benefits of multi-task deep neural networks, frequently showing outstanding performance when comparing with shallow models~\cite{ma2015deep,unterthiner2014deep,liu2018practical}.
All of these works used Morgan fingerprints as input representations.

Another option for molecule representation is the SMILES string~\cite{weininger1989smiles}.
SMILES can be treated as a sequence of atoms and bonds, and each molecule has a unique canonical SMILES string among a frequently vast set of noncanonical, but completely valid, SMILES strings.
Therefore, attempts were made to make SMILES feed into more complicated neural networks.
\cite{jastrzkebski2016learning} applied recurrent neural network language model (RNN) and convolutional neural networks (CNN) on SMILES, and showed that CNN is best when evaluated on the log-loss.
SMILES as the representation is now common in molecule generation tasks.
\cite{gomez2016automatic} first applied SMILES for automatic molecule design, and \cite{kusner2017grammar} proposed using a parser tree on SMILES so as to produce more grammatically-valid molecules, where the input is the one-hot encoded rules.
On the other hand, \cite{liu2018practical} showed the limitation of SMILES and itself as a structured data is hard to interpret, and thus SMILES are not used in our experiments.

Molecular descriptors~\cite{todeschini2009molecular} is another representation, but it requires heuristically coming up with descriptors and dynamically adjusting it to tasks, which is not easy and requires a lot of domain knowledge.
Therefore molecular descriptors are not considered in this paper since one of the goal here is to get a generalized feature representation.

Recent works started to explore the graph representation, and the benefit is its capability to encode the structured data.
\cite{duvenaud_convolutional_nodate} first utilized message passing on graphs.
At each step, this method passes the hidden message layer to the intermediate feature layer.
The summed-up neural fingerprints are then fed into neural networks as features.
Following this line of research, \cite{altae2017low} made small adaptations by using the last message layer as feature inputs for neural network, and \cite{ying2018hierarchical} proposed a differential pooling layer to learn the hierarchical information.

Other variants introduced different modules.
\cite{kearnes2016molecular} proposed a new module called weave for delivering information among atoms and bonds, and \cite{matlock2018learning} used a weave operation with forward and backward operations across a molecule graph.
\cite{li2015gated} utilized edge information, and \cite{faber2017prediction} generalized it into a message passing network framework, highlighting the importance of spatial information.


Viewing the molecules as graphs, the kernel method can be applied by using existing graph kernels (e.g., \cite{shawe2004kernel,shervashidze2011weisfeiler}). 
The implicit feature mapping induced by the kernel can be viewed as the representation for the input. The Weisfeiler-Lehman kernel~\cite{shervashidze2011weisfeiler} is particularly related due to its efficiency and theoretical backup. It is also similar in spirit to the Morgan fingerprints and closely related to the recent GIN graph neural network~\cite{xu2018powerful}.

\section{Background and Preliminaries} \label{s:background}


Generally, molecules can be viewed as graphs on atoms together with attribute information of the atoms, and we assume our molecule datasets are given in the format.\footnote{There can be other formats of raw data (such as 2D projections of the molecules), or missing data entries (such as missing attribute information for an atom). These are not considered here for simplicity. } 
To apply learning methods, they are converted to feature vectors (fingerprints), or are directly handled by specifically designed learning models (graph neural networks). The fingerprints or the hidden layers of graph neural networks are regarded as the representations or embeddings of the graphs.

\subsection{Raw Data: Representation as Graphs With Vertex Attributes} \label{s:raw_format}

Nearly all molecules can be potentially represented as a graph, where each atom is a vertex and each bond is an edge.
Suppose there are $m$ vertices in the graph, denoted as $i \in \{0, 1, ..., m-1\}$.
Each vertex entails useful attribute information, like the atom symbol and number of charges for atom vertices.
These vertex attributes are encoded into a vertex attribute matrix $\mathcal{V} \in \{0, 1\}^{m \times S}$, where $S$ is the number of attributes. A concrete example is given by the following: 
\begin{align} \label{eq:vertex_attribute_matrix}
    & \mathcal{V}_{i, \cdot} = [\mathcal{V}_{i,0}, \mathcal{V}_{i,1}, \hdots, \mathcal{V}_{i,6}, \mathcal{V}_{i,7}], 
    \nonumber\\
    & \text{atom symbol } \mathcal{V}_{i,0} \in \{ \text{C}, \text{Cl}, \text{I}, \text{F}, \hdots \},
    \nonumber\\
    & \text{atom degree } 
    \mathcal{V}_{i,1} \in \{
    0, 1, 2, 3, 4, 5, 6 \},
    \nonumber\\
    & \hdots
    \nonumber\\
    & \text{is acceptor } \mathcal{V}_{i,6} \in \{0,1\},
    \nonumber\\
    & \text{is donor } \mathcal{V}_{i,7} \in \{0,1\}.
\end{align}
Note that the attributes typically have discrete values. 

The bonding information is encoded into the adjacency matrix $\mathcal{A} \in \{0, 1\}^{m \times m}$, where  $\mathcal{A}_{i,j} = 1$ if and only if two vertices $i$ and $j$ are linked.

We let $G = (\mathcal{V}, \mathcal{A})$ denote a molecular graph. 

\subsection{Fingerprints}

We review two prototype methods here. Morgan fingerprints and its variants \cite{morgan1965generation} have been one of the most widely used featurization methods in virtual screening.
It is an iterative algorithm that encodes the circular substructures of the molecule as identifiers at increasing levels with each iteration.
In each iteration, hashing is applied to generate new identifiers, and thus, there is a chance that two substructures are represented by the same identifier.
In the end, a list of identifiers encoding the substructures is folded to bit positions of a fixed-length bit string.
A 1-bit at a particular position indicates the presence of a substructure (or multiple substructures if they are all hashed to this position) and a 0-bit indicates the absence of corresponding substructures.
Due to the hashing collisions, it is difficult to interpret such fingerprints and examine how the machine learning systems utilize them.

\begin{figure}[htb!]	
    \centering
    \includegraphics[width=0.85\linewidth]{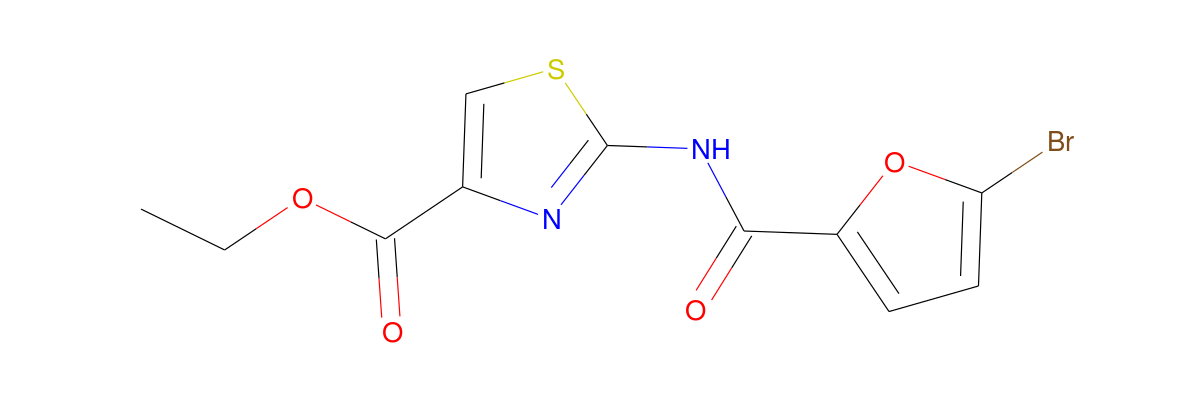}	
    \caption{Illustration of the Morgan fingerprint and SMILES molecule representations.	
             The molecule is displayed on a 2D space.	
             The corresponding canonical SMILES is c1cc(oc1C(=O)Nc2nc(cs2)C(=O)OCC)Br,	
             and Morgan Fingerprints is, for example, [000000...00100100100...000000]. }
    \label{fig:molecule_graph_representation}
\end{figure}

Another prototypical method, Simplified Molecular Input Line Entry System (SMILES)~\cite{weininger1989smiles}, is a character sequence describing molecular structures. There are some inherent issues in SMILES, the biggest being that molecules cannot be simply represented as a linear sequence: the properties of drug-like organic molecules usually have dependence on ring structures and tree-like branching, whose information is lost in a linear sequence. Our experiments show that it generally achieves worse performance than the other methods, so it is not considered as a competitor in the experimental section.

One example of molecule as a graph is shown in \Cref{fig:molecule_graph_representation}, together with its Morgan fingerprint and SMILES molecule representations.

\subsection{Graph Neural Networks} \label{sec:message_passing}

In recent works, message passing has been dominant in graph neural networks~\cite{pmlr-v70-gilmer17a,hamilton2017representation,zhou2018graph,wu2019comprehensive}.
A GNN keeps a vector $h_i$ for each vertex $i$ and uses some neighborhood aggregation strategy that iteratively updates the vector by aggregating those of its neighbors. After $t$ iterations, 
each vertex is able to capture the information of the vertices at most $t$-hops away. 
Formally, the $k$-th iteration is to compute
\begin{align} \label{eq:message_passing}
  f_i^{(k)} & = \text{AGGREGATE}^{(k)}(\{h_j^{(k-1)}: j \in \text{Neighbor}(i)\}), 
  \nonumber \\
  h^{(k)}_i & = \text{COMBINE}^{(k)}(h_i^{(k-1)}, f_i^{(k)}),
\end{align}
where $h^{(k)}_i$ is the value of $h_i$ at the $k$-th iteration, $h_i^{(0)}$ is typically initialized to the attribute vector of the vertex, and $\text{AGGREGATE}^{(k)}$ and $\text{COMBINE}^{(k)}$ are carefully chosen functions. The representation for the whole graph is then some aggregation of the vertex vectors. 
Such a framework has been used in the domains of molecules, but in general needs to be carefully specialized to this setting, see, e.g.,~\cite{duvenaud_convolutional_nodate,altae2017low,kearnes2016molecular}.

\section{Complete Proofs for Theoretical Analysis} 
\label{s:proof_analysis}

\subsection{Preliminary}
Here we provide a brief review of related concepts in the field of compressed sensing that are important for our analysis, following~\cite{arora2018acompressed,kasiviswanathan2019restricted}. For a review with details, please refer to~\cite{foucart2017mathematical}. 

The primary goal of compressed sensing is to recover a high-dimensional $k$-sparse signal $x \in \mathbb{R}^N$ from a few linear measurements. Here, being $k$-sparse means that $x$ has at most $k$ non-zero entries, i.e., $|x|_0 \le k$. In the noiseless case, we have a design matrix $A \in \mathbb{R}^{d \times N}$ and the measurement vector is $z = Ax$. The optimization formulation is then
\begin{align} \label{eqn:l0}
    \text{minimize}_{x'}  \|x'\|_0 \quad \text{subject to} \quad Ax' = z 
\end{align}
where $\|x'\|_0$ is $\ell_0$ norm of $x'$, i.e., the number of non-zero entries in $x'$. The assumption that $x$ is the sparsest vector satisfying $Ax = z$ is equivalent to that $x$ is the optimal solution for (\ref{eqn:l0}). 

Unfortunately, the $\ell_0$-minimization in (\ref{eqn:l0}) is NP-hard. The typical approach in compressed sensing is to consider its convex surrogate using $\ell_1$-minimization:
\begin{align} \label{eqn:l1}
    \text{minimize}_{x'}  \|x'\|_1 \quad \text{subject to} \quad Ax' = z 
\end{align}
where $\|x'\|_1 = \sum_i |x'_i|$ is the $\ell_1$ norm of $x'$. The fundamental question is when the optimal solution of (\ref{eqn:l0}) is equivalent to that of (\ref{eqn:l1}), i.e., when exact recovery is guaranteed. 

\subsubsection{The Restricted Isometry Property} 
One common condition for recovery is the Restricted Isometry Property (RIP):
\begin{definition}
$A \in \mathbb{R}^{d \times N}$ is $(\mathcal{X}, \epsilon)$-RIP for some subset $\mathcal{X} \subseteq \mathbb{R}^N$ if for any $x \in \mathcal{X}$,
\begin{align*}
    (1-\epsilon) \|x\|_2 \le \| A x\|_2 \le (1+\epsilon) \|x\|_2. 
\end{align*}
We will abuse notation and say $(k, \epsilon)$-RIP if $\mathcal{X}$ is the set of all $k$-sparse $x \in \mathbb{R}^N$.
\end{definition}
Introduced by~\cite{candes2005decoding}, RIP has been used to show to guarantee exact recovery. 
\begin{thm}[Restatement of Theorem 1.1 in \cite{candes2008restricted}]
\label{thm:recovery}
Suppose $A$ is $(2k, \epsilon)$-RIP for an $\epsilon < \sqrt{2} - 1$. Let $\hat{x}$ denote the solution to (\ref{eqn:l1}), and let $x_k$ denote the vector $x$ with all but the k-largest entries set to zero. Then
\begin{align*}
    \| \hat{x} - x \|_1 \le C_0 \|x_k - x\|_1
\end{align*}
and
\begin{align*}
    \| \hat{x} - x \|_2 \le C_0 k^{-1/2}  \|x_k - x\|_1.
\end{align*}
In particular, if $x$ is $k$-sparse, the recovery is exact. 
\end{thm}
Furthermore, it has been shown that $A$ is $(k,\epsilon)$-RIP with overwhelming probability when $d = \Omega(k \log \frac{N}{k})$ and $\sqrt{d} A_{ij} \sim \mathcal{N}(0,1) (\forall i,j)$ or $\sqrt{d} A_{ij} \sim \mathcal{U}\{-1, 1\} (\forall i,j)$. 

For our purpose, we also concern about whether the $\ell$-way column Hadamard-product of $A$ has RIP.
\begin{definition}[$\ell$-way Column Hadamard Product]
Let $A$ be a $d \times N$ matrix, and let $\ell$ be a natural integer. The $\ell$-way column Hadamard-product of $A$ is a $d \times {N \choose \ell}$ matrix denoted as $A^{(\ell)}$, whose columns indexed by a sequence $1 \le i_1 < i_2 \cdots < i_\ell \le d$ is the element-wise product of the $i_1, i_2, \dots, i_\ell$-th columns of $A$, i.e., $(i_1, i_2, \dots, i_\ell)$-th column in $A^{(\ell)}$ is $A_{i_1} \odot A_{i_2} \odot \cdots \odot A_{i_\ell}$ where $A_j$ for $j \in [N]$ is the $j$-th column in $A$.
\end{definition}
We have the following theorems:
\begin{thm}[Restatement of Theorem 4.1 in~\cite{kasiviswanathan2019restricted}]
\label{thm:hamadard}
Let $X$ be an $n \times d$ matrix, and let $A$ be a $d \times N$ random matrix with independent entries $R_{ij}$ such that $\mathbb{E}[R_{ij}] = 0, \mathbb{E}[R_{ij}] = 1$, and $|R_{ij}| \le \tau$ almost surely. 
Let $\epsilon  \in (0,1)$, and let $k$ be an integer satisfying $\text{sr}(X) \ge \frac{C \tau^{8}}{\epsilon^2} k^2 \log \frac{N^2}{k\epsilon}$ for some universal constant $C>0$. Then with probability at least $1 - \exp(-c \epsilon^2 \text{sr}(X) / (k^2 \tau^8))$ for some universal constant $c>0$, the matrix $X A^{(\ell)}/ \| X \|_F$ is $(k, \epsilon)$-RIP. 
\end{thm}
Here, $\text{sr}(X)= \| X\|_F^2 / \|X\|^2$ is the stable rank of $X$. In our case, we will apply the theorem with $X$ being $\mathbf{I}_{d \times d}/\sqrt{d}$ where $\mathbf{I}_{d \times d} \in \mathbb{R}^{d \times d}$ is the identity matrix. 

\begin{thm}[Restatement of Theorem 4.3 in~\cite{kasiviswanathan2019restricted}] \label{thm:hamadard_g}

Let $X$ be an $n \times d$ matrix, and let $A$ be a $d \times N$ random matrix with independent entries $R_{ij}$ such that $\mathbb{E}[R_{ij}] = 0, \mathbb{E}[R_{ij}] = 1$, and $|R_{ij}| \le \tau$ almost surely. Let $\ell \ge 3$ be a constant. 
Let $\epsilon  \in (0,1)$, and let $k$ be an integer satisfying $\text{sr}(X) \ge \frac{C \tau^{4\ell}}{\epsilon^2} k^3 \log \frac{N^\ell}{k\epsilon}$  for some universal constant $C>0$. Then with probability at least $1 - \exp(-c \epsilon^2 \text{sr}(X) / (k^2 \tau^{4 \ell}))$  for some universal constant $c>0$, the matrix $X A^{(\ell)} / \| X \|_F $ is $(k, \epsilon)$-RIP. 
\end{thm}

\subsubsection{Compressed Learning}
Given that $Ax$ preserves the information of sparse $x$ when $A$ is RIP, it is then natural to study the performance of a linear classifier learned on $Ax$ compared to that of the best linear classifier on $x$. Our analysis will use a theorem from~\cite{arora2018acompressed} that generalizes that of~\cite{calderbank2009compressed}. 

Let $\mathcal{X} \subseteq \mathbb{R}^N$ denote 
\begin{align*}
    \mathcal{X} = \{x: x \in \mathbb{R}^N, \|x\|_0 \le k, \|x\|_2 \le B\}. 
\end{align*}
Let $\{(x_i, y_i)\}_{i=1}^M$ be a set of $M$ samples i.i.d. from some distribution over $\mathcal{X} \times \{-1, 1\}$. Let $\ell$ denote a $\lambda_\ell$-Lipschitz convex loss function. Let $\ell_\mathcal{D}(\theta)$ denote the risk of a linear classifier with weight $\theta \in \mathbb{R}^N$, i.e., $\ell_\mathcal{D}(\theta) = \mathbb{E} [\ell( \langle \theta, x \rangle, y)]$, and let $\theta^*$ denote a minimizer of  $\ell_\mathcal{D}(\theta)$. Let $\ell^A_\mathcal{D}(\theta)$ denote the risk of a linear classifier with weight $\theta \in \mathbb{R}^d$ over $Ax$, 
i.e., $\ell^A_\mathcal{D}(\theta_A) = \mathbb{E} [\ell( \langle \theta_A, A x \rangle, y)]$, and let $\hat{\theta}_A$ denote the weight learned with $\ell_2$-regularization over $\{(Ax_i, y_i)\}_i$:
\begin{align}
\hat{\theta}_A = \arg\min_{\theta}  \frac{1}{M} \sum_{i=1}^M \ell(\langle \theta, Ax_i \rangle, y_i)  + \lambda \| \theta \|_2  
\end{align}
where $\lambda$ is the regularization coefficient. 

\begin{thm}
[Restatement of Theorem 4.2 in~\cite{arora2018acompressed}]
\label{thm:compressedlearning}
Suppose $A$ is $(\Delta \mathcal{X}, \epsilon)$-RIP.
Then with probability at least $1-\delta$, 
\begin{align*}
    \ell^A_\mathcal{D}(\hat{\theta}_A) 
    \le 
    \ell_\mathcal{D}(\theta^*) + O\left( \lambda_\ell B \|\theta^*\| \sqrt{\epsilon + \frac{1}{M} \log \frac{1}{\delta} } \right)
\end{align*}
for appropriate choice of $C$. Here, $\Delta \mathcal{X} = \{x  - x': x, x' \in \mathcal{X}\}$ for any $\mathcal{X} \subseteq \mathbb{R}^N$. 
\end{thm}

\subsection{Representation Power}
In this subsection, we provide the proof of Theorem~\ref{thm:rep}.

We begin by defining the distribution over the vertex embedding matrix $W$. Recall that $k_j$ is the number of possible values for the $j$-th attribute. Suppose we have numbers $r_j \in (0, r)$  so that $\sum_{j=0}^{S-1} r_j = r$ whose values will be specified later. Let 
\begin{align}
W = \begin{bmatrix}
U^0 & 0 & \cdots & 0 \\
0   & U^1 & \cdots & 0 \\
\cdots & \cdots & \cdots & \cdots \\
0  &  0 & \cdots & U^{S-1}
\end{bmatrix}
\end{align}
where $U^j \in \mathbb{R}^{r_j \times k_j}$. 
Now let's specify $U^j$. 
Let the entries in $U^j$'s are independent random variables, and let the entries be uniform from $\{-1, 1\}$ with some scaling factor $c_u$, i.e., $(U^j)_{ik} \sim c_u \times \mathcal{U}\{-1, 1\}$, where the value of $c_u$ will be determined later.%
\footnote{
In fact, the entries can be $c$ times any distribution that has mean 0, variance 1, and is almost surely bounded by a constant.
}

Now, let $(U^j)^{(n)}$ denote the $n$-way column Hadamard product of $U^j$, and let
\begin{align} \label{eqn:Tpn}
T_{(n)} = \begin{bmatrix}
(U^0)^{(n)} & 0 & \cdots & 0 \\
0   & (U^1)^{(n)} & \cdots & 0 \\
\cdots & \cdots & \cdots & \cdots \\
0  &  0 & \cdots & (U^{S-1})^{(n)}
\end{bmatrix}
\end{align}
Then it can be verified that 
\begin{align}
    f_{(n)} = T_{(n)} c_{(n)}.
\end{align}
Now we can apply Theorem~\ref{thm:hamadard} for $n=2$ and Theorem~\ref{thm:hamadard_g} for $n\ge 3$ on each $(U^j)^{(n)}$. Let $s_{n,j}$ denote the sparsity of $c^{(j)}_{(n)}$. Then with $r_j \ge \Omega(n s_{n,j}^3  \log k_j)$ and appropriate set scaling factor $c_u$, we have that with probability at least $1-\exp(-c r_j /s_{n,j}^2)$, $(U^j)^{(n)}$ is $(2s_{n,j},\epsilon)$-RIP for $\epsilon=0.1$. This then means that $f_{n}$ can be exactly recovered from $c_{(n)}$ by Theorem~\ref{thm:recovery}.
Now, by setting $r = \Omega(n s_n^3 \log K)$ where $s_n = \sum_{j=0}^{S-1} s_{n,j}$, we can choose $r_j$'s satisfying $r_j = \Omega(n s_{n,j}^3  \log k_j + r/S)$. Furthermore, we have $r_j /s_{n,j}^2 = \Omega(r_j^{1/3}) = \Omega((r/S)^{1/3})$, so the failure probability is bounded by 
$ S \exp(-c (r/S)^{1/3})$.

\subsection{Prediction Power}

\paragraph{Proof of Theorem~\ref{thm:pre}.}
Theorem~\ref{thm:pre} is a direct consequence of Theorem~\ref{thm:rep}.

By Theorem~\ref{thm:rep}, under the conditions, we have that there exists a mapping $\mathcal{M}_{(n)}$ from $f_G$ to $c_{(n)}$. Therefore, there exists a mapping $\mathcal{M}_{[T]}$ from $f_G$ to $c_{[T]}$, by applying $\mathcal{M}_{(n)}$'s on each blocks of $f_G$, respectively.
Now, define $g_f = g_c \circ \mathcal{M}_{[T]}$, such that $g_f(f_G) = g_c \circ \mathcal{M}_{[T]}(f_G) = g_c(c_{[T]})$, so $\ell_\mathcal{D}(g_f) = \ell_\mathcal{D}(g_c)$.

\paragraph{Proof of Theorem~\ref{thm:pre_linear}}
Let 
\begin{align}
T_{[T]} = \begin{bmatrix}
T_{(1)} & 0 & \cdots & 0 \\
0   & T_{(2)} & \cdots & 0 \\
\cdots & \cdots & \cdots & \cdots \\
0  &  0 & \cdots & T_{(T)}
\end{bmatrix}
\end{align}
where $T_{(n)} (1\le n \le T)$ is defined as in (\ref{eqn:Tpn}).
Then it can be verified that 
\begin{align}
    f_{G} = T_{[T]} c_{[T]}.
\end{align}
Under the specified conditions we have that with high probability $T_{(n)}$'s are  $(2s_n, \epsilon)$-RIP, so $T_{[T]}$ is $(\Delta \mathcal{X}, \epsilon)$-RIP. Since the logistic loss is 1-Lipschitz convex, the statement follows from Theorem~\ref{thm:compressedlearning}, while the failure probability follows from a union bound.

\section{Task Specification} \label{s:task_specification}
\begin{table}[H]
    \centering
    \small
    \caption{
        Number of positives and all molecules on 12 Tox21 tasks.
    }
    \begin{tabular}{|C{3cm}|c|c|c|}
        \hline
        Task & Num of Positives & Total Number & Positive Ratio (\%)\\
        \hline
        \hline
        NR-AR&304&7332&4.14621 \\
        NR-AR-LBD&237&6817&3.47660 \\
        NR-AhR&783&6592&11.87803 \\
        NR-Aromatase&298&5853&5.09141 \\
        NR-ER&784&6237&12.57015 \\
        NR-ER-LBD&347&7014&4.94725 \\
        NR-PPAR-gamma&186&6505&2.85934 \\
        SR-ARE&954&5907&16.15033 \\
        SR-ATAD5&262&7140&3.66947 \\
        SR-HSE&378&6562&5.76044 \\
        SR-MMP&912&5834&15.63250 \\
        SR-p53&414&6814&6.07573 \\
        \hline
    \end{tabular}
    \label{tab:tox21_distribution}
\end{table}

\begin{table}[H]
    \centering
    \small
    \caption{
        Number of positives and all molecules on 2 ClinTox tasks
    }
    \begin{tabular}{|C{3cm}|c|c|c|}
        \hline
        Task & Num of Positives & Total Number & Positive Ratio (\%)\\
        \hline
        \hline
        CT\_TOX&112&1469&7.62423 \\
        FDA\_APPROVED&1375&1469&93.60109 \\
        \hline
    \end{tabular}
    \label{tab:clintox_distribution}
\end{table}

\begin{table}[H]
    \centering
    \small
    \caption{
        Number of positives and all molecules on 17 MUV tasks.
    }
    \begin{tabular}{|C{3cm}|c|c|c|}
        \hline
        Task & Num of Positives & Total Number & Positive Ratio (\%)\\
        \hline
        \hline
        MUV-466&27&14844&0.18189 \\
        MUV-548&29&14737&0.19678 \\
        MUV-600&30&14734&0.20361 \\
        MUV-644&30&14633&0.20502 \\
        MUV-652&29&14903&0.19459 \\
        MUV-689&29&14606&0.19855 \\
        MUV-692&30&14647&0.20482 \\
        MUV-712&28&14415&0.19424 \\
        MUV-713&29&14841&0.19540 \\
        MUV-733&28&14691&0.19059 \\
        MUV-737&29&14696&0.19733 \\
        MUV-810&29&14646&0.19801 \\
        MUV-832&30&14676&0.20442 \\
        MUV-846&30&14714&0.20389 \\
        MUV-852&29&14658&0.19784 \\
        MUV-858&29&14775&0.19628 \\
        MUV-859&24&14751&0.16270 \\
        \hline
    \end{tabular}
    \label{tab:muv_distribution}
\end{table}

\begin{table}[H]
    \centering
    \small
    \caption{
        Number of positives and all molecules on 1 HIV task.
    }
    \begin{tabular}{|C{3cm}|c|c|c|}
        \hline
        Task & Num of Positives & Total Number & Positive Ratio (\%)\\
        \hline
        \hline
        HIV&1425&41023&3.47366 \\
        \hline
    \end{tabular}
    \label{tab:hiv_distribution}
\end{table}

\section{Atom Feature Specification} \label{s:attribute_specification}
\Cref{tab:vertex_attribute_matrix,tab:vertex_attribute_matrix_exception} show the types of feature attributes for the atoms in the molecules of the datasets used in our experiments.
Also in \Cref{sec:feature_effects}, we can observe that the selection of feature attribute values, especially adding more atom symbols, has very limited improvement.
\begin{table}[htb!]
    \small
    \centering
    \caption{$d=42$ features are divided into $S=8$ attributes.
             Each feature attribute corresponds to one type of atom property, including atom symbol, atom degree, atom charge, etc.
             Note that it is hard to enumerate all the values for the atom properties, so we use the last bit 'Unknown' as the placeholder to catch the missing symbols.
}
    \label{tab:vertex_attribute_matrix}
    \begin{tabular}{|C{0.5cm}|C{1cm}|C{3cm}|C{6cm}|}
        \hline
        id & digit & property & values\\
        \hline
        \hline
        0 & 0-9 & atom symbol & [C, Cl, I, F, O, N, P, S, Br, Unknown]\\
        1 & 10-16 & atom degree & [0, 1, 2, 3, 4, 5, Unknown]\\
        2 & 17-23 & number of Hydrogen & [0, 1, 2, 3, 4, 5, Unknown]\\
        3 & 24-29 & implicit valence & [0, 1, 2, 3, 4, Unknown]\\
        4 & 30-35 & atom charge & [-2, -1, 0, 1, 2, Unknown]\\
        5 & 36-37 & is aromatic & [no, yes]\\
        6 & 38-39 & is acceptor & [no, yes]\\
        7 & 40-41 & is donor & [no, yes]\\
        \hline
    \end{tabular}
\end{table}

\begin{table}[htb!]
    \centering
    \caption{
        On datasets QM8 and QM9, due to the input format of the molecule file, we cannot extract the atom attributes like the number of hydrogen, is-acceptor and is-donor property (while keeping its 3D information at the same time).
        So following \cite{wu2018moleculenet}, only $d=32$ features, \ie, $S=5$ attributes are considered.
    }
    \label{tab:vertex_attribute_matrix_exception}
    \begin{tabular}{|C{0.5cm}|C{1cm}|C{3cm}|C{6cm}|}
        \hline
        id & digit & property & values\\
        \hline
        \hline
        0 & 0-9 & atom symbol & [C, Cl, I, F, O, N, P, S, Br, Unknown]\\
        1 & 10-16 & atom degree & [0, 1, 2, 3, 4, 5, Unknown]\\
        2 & 17-23 & implicit valence & [0, 1, 2, 3, 4, 5, Unknown]\\
        3 & 24-29 & atom charge & [-2, -1, 0, 1, 2, Unknown]\\
        4 & 30-31 & is aromatic & [no, yes]\\
        \hline
    \end{tabular}
\end{table}

\newpage

\section{Hyperparameter Tuning} \label{s:hyperparameter}

\subsection{Hyperparameters for Representation}

\paragraph{Morgan Fingerprints.}
To generate Morgan fingerprints, we use the public package RdKit \cite{Landrum2016RDKit2016_09_4} and follow the hyperparameters from benchmark \cite{Ramsundar-et-al-2019}: the number of bits is 1024 and radius is 2.

\paragraph{Graph Neural Networks.}
For Graph CNN, Weave Neural Network, Deep Tensor Neural Network, and Message-Passing Neural Network, we follow the optimal hyperparameter schemes provided in \cite{wu2018moleculenet}. Note that they are tuned for each of these datasets, respectively, to guarantee the optimality.

\paragraph{N-Gram Graph.}
The hyperparameters for N-gram graph are included in \Cref{tab:hyper_n_gram_graph}, and the effects of two important hyperparameters (random dimension $r$ and n-gram number $T$) will be discussed in \Cref{s:effect_hyper}.

\begin{table}[H]
	\centering
	\small
    \caption{Hyperparameter sweeping for N-Gram Graph. We have $S=8$ feature attributes.}
    \label{tab:hyper_n_gram_graph}
    \begin{tabular}{|C{4cm}|C{8cm}|}
    \hline
	Hyperparameters & Candidate values \\
    \hline
    \hline
    Random Dimension $r$ & 50, 100\\
    N-Gram Num $T$ & 2, 4, 6\\
    Embedding Structure & [Embedding -> Sum], [Embedding -> Mean]\\
    Neural Network & [$r$, 20, $S$], [$r$, 100, $S$] , [$r$, 100, 20, $S$]\\
	\hline
    \end{tabular}
\end{table}

\subsection{Hyperparameters for Modeling}

For other baseline models, we run a grid search for hyperparameter sweeping, including Weisfeiler-Lehman Graph Kernel in \Cref{tab:hyper_wl}, random forest in \Cref{tab:hyper_rf}, XGBoost in \Cref{tab:hyper_xgb}, and Graph Isomorphism Network \Cref{tab:hyper_gin}.

\begin{table}[H]
    \centering
	\small
    \caption{Hyperparameter sweeping for Weisfeiler-Lehman Graph Kernel.}
    \label{tab:hyper_wl}
    \begin{tabular}{|C{4cm}|C{8cm}|}
    \hline
	Hyperparameters & Candidate values \\
	\hline
	\hline
	Number of Step & 1, 2, 3 \\
    \hline
    \end{tabular}
\end{table}

\begin{table}[H]
	\centering
	\small
    \caption{Hyperparameter sweeping for Random Forest.}
    \label{tab:hyper_rf}
    \begin{tabular}{|C{4cm}|C{8cm}|}
    \hline
	Hyperparameters & Candidate values \\
    \hline
    \hline
    Number of Trees	&	100, 4000 \\
    Max Features	&	None, sqrt, log2 \\
    Min Samples Leaf	&	1, 10, 100, 1000 \\
    Class Weight	&	None, balanced\_subsample, balanced \\
	\hline
    \end{tabular}
\end{table}

\begin{table}[H]
	\centering
	\small
    \caption{Hyperparameter sweeping for XGBoost.}
    \label{tab:hyper_xgb}
    \begin{tabular}{|C{4cm}|C{8cm}|}
    \hline
	Hyperparameters & Candidate values \\
    \hline
    \hline
    Max Depth	&	5, 10, 50, 100 \\
    Learning Rate	&	1, 3e-1, 1e-1, 3e-2 \\
    Number of Trees	&	30, 100, 300, 1000, 3000 \\
	\hline
    \end{tabular}
\end{table}

\begin{table}[H]
	\centering
	\small
    \caption{Hyperparameter sweeping for Graph Isomorphism Network.}
    \label{tab:hyper_gin}
    \begin{tabular}{|C{4cm}|C{8cm}|}
    \hline
	Hyperparameters & Candidate values \\
    \hline
    \hline
    Max Depth	&	2, 3, 5 \\
    Hidden Dimension & 30, 50 \\
    Epoch & 100, 300 \\
    Optimizer & SGD, Adam \\
    Learning Rate Scheduler & None, ReduceLROnPlateau, StepLR\\
	\hline
    \end{tabular}
\end{table}

\newpage

\section{Vertex Embedding} \label{s:vertex_embedding}

The CBoW-like neural network structure is displayed in \Cref{fig:cbow_projection}.
Though the vertex embedding step is unsupervised, we still follow the 5-fold cross-validation, so as not to touch the test set before prediction. In other words, we will create 5 CBoW-like models for each task (or dataset \footnote{For regression tasks like QM8, QM9, and Clintox, all the molecules are sharing the same splits since they don't have any restrictions like missing labels or stratified splits.}) and each vertex embedding dimension $R$. We report the test accuracy during vertex embedding in \Cref{tab:vertex_embedding_acc}.

\vspace{-2mm}

\begin{table}[H]
	\small
    \centering
    \caption{
        The mean accuracy of 5-fold cross-validation in vertex embedding.
        The accuracy measures how the CBoW-like neural network can accurately predict the vertex attributes.
    }
    \label{tab:vertex_embedding_acc}
    \begin{tabular}{|C{3cm}|C{4cm}|C{4cm}|}
        \hline
        Task/Dataset & Accuracy(\%), $r=50$ & Accuracy(\%), $r=100$\\
        \hline
        Delaney&$0.924 \pm 0.001$&$0.924 \pm 0.001$\\
        Malaria&$0.937 \pm 0.001$&$0.938 \pm 0.000$\\
        CEP&$0.923 \pm 0.001$&$0.923 \pm 0.000$\\
        QM7&$0.898 \pm 0.001$&$0.897 \pm 0.001$\\
        QM8&$0.988 \pm 0.000$&$0.988 \pm 0.000$\\
        QM9&$0.987 \pm 0.000$&$0.987 \pm 0.001$\\
        NR-AR&$0.916 \pm 0.000$&$0.916 \pm 0.000$\\
        NR-AR-LBD&$0.915 \pm 0.001$&$0.914 \pm 0.003$\\
        NR-AhR&$0.916 \pm 0.000$&$0.915 \pm 0.000$\\
        NR-Aromatase&$0.915 \pm 0.000$&$0.915 \pm 0.000$\\
        NR-ER&$0.915 \pm 0.001$&$0.914 \pm 0.001$\\
        NR-ER-LBD&$0.916 \pm 0.001$&$0.915 \pm 0.001$\\
        NR-PPAR-gamma&$0.914 \pm 0.000$&$0.915 \pm 0.000$\\
        SR-ARE&$0.915 \pm 0.000$&$0.915 \pm 0.000$\\
        SR-ATAD5&$0.916 \pm 0.000$&$0.916 \pm 0.000$\\
        SR-HSE&$0.915 \pm 0.000$&$0.915 \pm 0.000$\\
        SR-MMP&$0.913 \pm 0.001$&$0.914 \pm 0.001$\\
        SR-p53&$0.915 \pm 0.000$&$0.915 \pm 0.001$\\
        Clintox&$0.911 \pm 0.001$&$0.911 \pm 0.000$\\
        MUV-466&$0.940 \pm 0.001$&$0.940 \pm 0.000$\\
        MUV-548&$0.932 \pm 0.001$&$0.931 \pm 0.001$\\
        MUV-600&$0.938 \pm 0.000$&$0.937 \pm 0.001$\\
        MUV-644&$0.932 \pm 0.001$&$0.932 \pm 0.000$\\
        MUV-652&$0.939 \pm 0.000$&$0.939 \pm 0.001$\\
        MUV-689&$0.942 \pm 0.001$&$0.942 \pm 0.001$\\
        MUV-692&$0.934 \pm 0.001$&$0.934 \pm 0.000$\\
        MUV-712&$0.938 \pm 0.001$&$0.938 \pm 0.001$\\
        MUV-713&$0.938 \pm 0.000$&$0.937 \pm 0.000$\\
        MUV-733&$0.937 \pm 0.001$&$0.937 \pm 0.001$\\
        MUV-737&$0.940 \pm 0.001$&$0.939 \pm 0.001$\\
        MUV-810&$0.937 \pm 0.000$&$0.937 \pm 0.000$\\
        MUV-832&$0.937 \pm 0.001$&$0.936 \pm 0.001$\\
        MUV-846&$0.938 \pm 0.000$&$0.938 \pm 0.000$\\
        MUV-852&$0.937 \pm 0.001$&$0.937 \pm 0.001$\\
        MUV-858&$0.938 \pm 0.001$&$0.938 \pm 0.001$\\
        MUV-859&$0.939 \pm 0.001$&$0.939 \pm 0.001$\\
        HIV&$0.920 \pm 0.001$&$0.919 \pm 0.001$\\
    \hline
    \end{tabular}
\end{table}

\vspace{-2mm}

\subsection{Transferable Vertex Embedding} \label{s:transferable_vertex_embedding}

The complete process for getting \Cref{tab:transferrable} is as follows.

\noindent\textbf{Vertex Embedding.}
Train the unsupervised CBoW model for vertex embedding $W$ on all the molecules from the source dataset. For random projection, we just initialize parameters of the CBoW model under the Gaussian distribution, and only molecules for that task is used if it comes from Tox21, \ie, the non-transfer case.

\noindent\textbf{Graph Embedding.}
Apply $W$ on molecules from target task for the graph embedding, $f_G$. Then train the model based on $f_G$.

\newpage

\section{Complete Results on 60 Regression and Classification Tasks} \label{s:complete_performance}
\begin{table}[H]
	\scriptsize
    \centering
    \caption{
        Here we include the performance on 28 regression tasks with 7 models.
        All experiments are done on a 5-fold cross-validation, and the mean evaluation of 5 runs is reported here.
        The top-3 models are \textbf{bolded}, and the best model is \underline{\textbf{underlined}}.
    }
    \label{tab:complete_performance_regression}
    \begin{tabular}{|c|c|c|c|c|c|c|c|c|}
\hline
Task&Eval Metric&\makecell{WL\\SVM}&\makecell{Morgan\\RF}&\makecell{Morgan\\XGB}&GCNN&Weave&\makecell{N-Gram\\RF}&\makecell{N-Gram\\XGB}
\\
\hline
\hline
Delaney&RMSE
&1.265 &1.168 &3.063 &0.825 &\underline{\textbf{0.687}} &\textbf{0.769} &\textbf{0.731} \\
\hline
Malaria&RMSE
&1.094 &\underline{\textbf{0.983}} &1.943 &1.144 &1.487 &\textbf{1.022} &\textbf{1.019} \\
\hline
CEP&RMSE
&1.800 &\underline{\textbf{1.300}} &3.049 &1.493 &2.846 &\textbf{1.399} &\textbf{1.366} \\
\hline
QM7&MAE
&176.750 &127.662 &110.230 &76.637 &\textbf{62.560} &\textbf{57.747} &\underline{\textbf{53.919}} \\
\hline
E1-CC2&MAE
&0.032 &0.008 &0.008 &\underline{\textbf{0.006}} &\textbf{0.007} &0.008 &\textbf{0.007} \\
E2-CC2&MAE
&0.023 &0.010 &0.010 &\textbf{0.008} &\underline{\textbf{0.007}} &0.009 &\textbf{0.008} \\
f1-CC2&MAE
&0.072 &\underline{\textbf{0.014}} &\textbf{0.015} &\textbf{0.014} &0.018 &0.015 &0.015 \\
f2-CC2&MAE
&0.081 &\textbf{0.032} &0.033 &\textbf{0.031} &0.036 &0.033 &\underline{\textbf{0.031}} \\
E1-PBE0&MAE
&0.034 &0.008 &0.008 &\textbf{0.006} &\underline{\textbf{0.006}} &0.008 &\textbf{0.007} \\
E2-PBE0&MAE
&0.029 &0.010 &0.010 &\underline{\textbf{0.007}} &\textbf{0.008} &0.008 &\textbf{0.008} \\
f1-PBE0&MAE
&0.068 &\textbf{0.012} &0.013 &\underline{\textbf{0.012}} &0.014 &0.013 &\textbf{0.013} \\
f2-PBE0&MAE
&0.078 &0.026 &0.027 &\textbf{0.024} &0.027 &\textbf{0.025} &\underline{\textbf{0.024}} \\
E1-CAM&MAE
&0.033 &0.007 &0.007 &\underline{\textbf{0.006}} &\textbf{0.006} &0.007 &\textbf{0.007} \\
E2-CAM&MAE
&0.025 &0.009 &0.009 &\underline{\textbf{0.006}} &\textbf{0.006} &0.008 &\textbf{0.007} \\
f1-CAM&MAE
&0.073 &\textbf{0.013} &0.014 &\underline{\textbf{0.013}} &0.016 &0.014 &\textbf{0.014} \\
f2-CAM&MAE
&0.080 &0.028 &0.028 &\underline{\textbf{0.026}} &0.031 &\textbf{0.028} &\textbf{0.026} \\
average& &0.052 &0.015 &0.015 &0.013 &0.015 &0.015 &0.014 \\
\hline
mu&MAE
&-- &0.548 &\textbf{0.533} &\underline{\textbf{0.482}} &0.624 &0.562 &\textbf{0.535} \\
alpha&MAE
&-- &3.787 &2.672 &\textbf{0.685} &1.034 &\textbf{0.722} &\underline{\textbf{0.612}} \\
homo&MAE
&-- &0.006 &0.006 &\underline{\textbf{0.004}} &\textbf{0.005} &0.005 &\textbf{0.005} \\
lumo&MAE
&-- &0.007 &0.006 &\underline{\textbf{0.004}} &\textbf{0.005} &0.006 &\textbf{0.005} \\
gap&MAE
&-- &0.008 &0.008 &\underline{\textbf{0.006}} &0.008 &\textbf{0.007} &\textbf{0.007} \\
r2&MAE
&-- &94.815 &82.516 &\textbf{64.775} &\underline{\textbf{42.095}} &72.846 &\textbf{59.137} \\
zpve&MAE
&-- &0.009 &0.007 &\textbf{0.001} &0.002 &\textbf{0.001} &\underline{\textbf{0.000}} \\
cv&MAE
&-- &1.505 &1.166 &0.524 &\textbf{0.374} &\textbf{0.434} &\underline{\textbf{0.334}} \\
u0&MAE
&-- &16.410 &12.736 &2.460 &\textbf{1.465} &\textbf{0.429} &\underline{\textbf{0.427}} \\
u298&MAE
&-- &16.410 &12.757 &2.671 &\textbf{1.560} &\textbf{0.429} &\underline{\textbf{0.428}} \\
h298&MAE
&-- &16.411 &12.752 &2.542 &\textbf{1.414} &\textbf{0.428} &\underline{\textbf{0.428}} \\
g298&MAE
&-- &16.414 &12.750 &2.466 &\textbf{2.359} &\textbf{0.428} &\underline{\textbf{0.428}} \\
average& &-- &13.823 &11.476 &6.474 &4.187 &6.357 &5.152 \\
\hline
    \end{tabular}
\end{table}

\newpage
\begin{table}[htb!]
    \scriptsize
    \centering
    \caption{
        Here we include the performance on 32 classification tasks with 8 models.
        All experiments are done on a 5-fold cross-validation, and the mean evaluation of 5 runs is reported here.
        The top-3 models are \textbf{bolded}, and the best model is \underline{\textbf{underlined}}.
    }
    \label{tab:complete_performance_classification}
    \begin{tabular}{|c|c|c|c|c|c|c|c|c|c|}
\hline
Task&Eval Metric&\makecell{WL\\SVM}&\makecell{Morgan\\RF}&\makecell{Morgan\\XGB}&GCNN&Weave&GIN&\makecell{N-Gram\\RF}&\makecell{N-Gram\\XGB}
\\
\hline
\hline
NR-AR&ROC-AUC
&0.759 &0.781 &0.780 &\textbf{0.793} &0.789 &0.755 &\underline{\textbf{0.797}} &\textbf{0.791} \\
NR-AR-LBD&ROC-AUC
&0.843 &\textbf{0.868} &0.853 &0.851 &0.835 &0.826 &\underline{\textbf{0.871}} &\textbf{0.864} \\
NR-AhR&ROC-AUC
&0.879 &\textbf{0.900} &0.894 &0.894 &0.870 &0.880 &\textbf{0.894} &\underline{\textbf{0.902}} \\
NR-Aromatase&ROC-AUC
&\textbf{0.849} &0.828 &0.780 &0.839 &0.819 &0.818 &\textbf{0.858} &\underline{\textbf{0.869}} \\
NR-ER&ROC-AUC
&0.716 &\textbf{0.731} &0.722 &0.731 &0.712 &0.688 &\textbf{0.747} &\underline{\textbf{0.753}} \\
NR-ER-LBD&ROC-AUC
&0.794 &0.806 &0.795 &0.806 &\textbf{0.808} &0.778 &\textbf{0.827} &\underline{\textbf{0.838}} \\
NR-PPAR-gamma&ROC-AUC
&0.819 &\textbf{0.844} &0.805 &0.817 &0.794 &0.800 &\underline{\textbf{0.856}} &\textbf{0.851} \\
SR-ARE&ROC-AUC
&0.803 &\textbf{0.814} &0.800 &0.799 &0.771 &0.788 &\textbf{0.826} &\underline{\textbf{0.835}} \\
SR-ATAD5&ROC-AUC
&0.819 &\textbf{0.854} &0.829 &0.825 &0.778 &0.814 &\textbf{0.857} &\underline{\textbf{0.860}} \\
SR-HSE&ROC-AUC
&\textbf{0.798} &0.782 &0.770 &0.772 &0.751 &0.723 &\textbf{0.798} &\underline{\textbf{0.812}} \\
SR-MMP&ROC-AUC
&0.887 &0.886 &0.878 &\textbf{0.894} &0.887 &0.866 &\textbf{0.911} &\underline{\textbf{0.918}} \\
SR-p53&ROC-AUC
&0.835 &\textbf{0.859} &0.796 &0.834 &0.795 &0.819 &\textbf{0.859} &\underline{\textbf{0.868}} \\
average& &0.813 &0.827 &0.806 &0.821 &0.798 &0.791 &0.841 &0.842 \\
\hline
CT\_TOX&ROC-AUC
&0.837 &0.788 &0.840 &\textbf{0.872} &\textbf{0.859} &0.823 &0.857 &\underline{\textbf{0.873}} \\
FDA\_APPROVED&ROC-AUC
&\textbf{0.851} &0.784 &0.830 &\underline{\textbf{0.875}} &0.836 &0.848 &0.825 &\textbf{0.874} \\
average& &0.833 &0.787 &0.847 &0.868 &0.834 &0.837 &0.837 &0.870 \\
\hline
MUV-466&PR-AUC
&0.046 &\textbf{0.076} &0.058 &0.003 &0.017 &\textbf{0.060} &0.058 &\underline{\textbf{0.086}} \\
MUV-548&PR-AUC
&\textbf{0.178} &\textbf{0.230} &\underline{\textbf{0.259}} &0.065 &0.065 &0.070 &0.073 &0.094 \\
MUV-600&PR-AUC
&\underline{\textbf{0.023}} &\textbf{0.021} &\textbf{0.017} &0.004 &0.006 &0.013 &0.007 &0.009 \\
MUV-644&PR-AUC
&\textbf{0.149} &\textbf{0.185} &\underline{\textbf{0.225}} &0.034 &0.025 &0.124 &0.046 &0.064 \\
MUV-652&PR-AUC
&\underline{\textbf{0.164}} &\textbf{0.095} &0.039 &0.020 &0.021 &0.022 &0.085 &\textbf{0.118} \\
MUV-689&PR-AUC
&\textbf{0.030} &0.025 &\underline{\textbf{0.094}} &0.011 &0.013 &0.021 &0.026 &\textbf{0.046} \\
MUV-692&PR-AUC
&0.003 &\underline{\textbf{0.010}} &0.003 &0.004 &0.003 &\textbf{0.006} &\textbf{0.005} &0.005 \\
MUV-712&PR-AUC
&\underline{\textbf{0.208}} &0.119 &\textbf{0.158} &0.062 &0.075 &\textbf{0.192} &0.134 &0.151 \\
MUV-713&PR-AUC
&\textbf{0.036} &\underline{\textbf{0.057}} &0.024 &0.007 &0.011 &0.007 &\textbf{0.026} &0.026 \\
MUV-733&PR-AUC
&\textbf{0.076} &\underline{\textbf{0.080}} &0.046 &0.011 &0.005 &0.016 &0.021 &\textbf{0.047} \\
MUV-737&PR-AUC
&0.058 &0.056 &\textbf{0.060} &0.008 &0.017 &0.005 &\underline{\textbf{0.084}} &\textbf{0.080} \\
MUV-810&PR-AUC
&\textbf{0.139} &\textbf{0.186} &\underline{\textbf{0.215}} &0.010 &0.006 &0.033 &0.013 &0.022 \\
MUV-832&PR-AUC
&0.365 &\underline{\textbf{0.556}} &\textbf{0.508} &0.029 &0.032 &\textbf{0.388} &0.229 &0.280 \\
MUV-846&PR-AUC
&\textbf{0.369} &0.299 &\underline{\textbf{0.407}} &0.219 &0.250 &\textbf{0.397} &0.250 &0.220 \\
MUV-852&PR-AUC
&\underline{\textbf{0.405}} &0.173 &\textbf{0.300} &0.159 &0.131 &\textbf{0.337} &0.214 &0.238 \\
MUV-858&PR-AUC
&\textbf{0.079} &\underline{\textbf{0.090}} &0.018 &0.006 &0.003 &\textbf{0.051} &0.014 &0.015 \\
MUV-859&PR-AUC
&0.004 &0.004 &\textbf{0.007} &0.005 &0.004 &0.003 &\underline{\textbf{0.007}} &\textbf{0.006} \\
average& &0.154 &0.146 &0.158 &0.041 &0.047 &0.119 &0.088 &0.099 \\
\hline
HIV&ROC-AUC
&0.800 &\underline{\textbf{0.849}} &0.827 &0.805 &0.663 &0.785 &\textbf{0.828} &\textbf{0.830} \\
\hline
    \end{tabular}
\end{table}

\newpage
\section{N-Gram Walk vs. N-Gram Path}
\label{s:walk_path_comparison}
We compare N-Gram Path (the version of N-gram graph method that excludes walks containing two vertices with the same attribute values) and N-Gram Walk (the version of N-gram graph method that does not exclude such walks) on each of the 60 tasks.
The same vertex embeddings are used, and both random forest (RF) and XGBoost (XGB) are experimented on top of the N-Gram graph embeddings.
All regression tasks are shown in \Cref{tab:walk_path_regression_results} and all classification tasks are shown in \Cref{tab:walk_path_classification_results}, and we can see that N-Gram Path is comparable to N-Gram Walk.

\begin{table}[H]
    \small
    \centering
    \caption{
        Comparison of N-Gram Path-based graph and N-Gram Walk-based graph on 28 regression tasks.
        All experiments are done on a 5-fold cross-validation, and the mean evaluation of 5 runs is reported here.
        N-Gram Walk-based graph with XGB trained on top of it can excel on 27 out of the 28 regression tasks, while N-Gram Path achieves comparable performance. 
    }
    \label{tab:walk_path_regression_results}
    \begin{tabular}{|C{2.5cm}|c|c|c|c|c|}
\hline
Task&Eval Metric&\makecell{N-Gram, RF\\Path}&\makecell{N-Gram, XGB\\Path}&\makecell{N-Gram, RF\\Walk}&\makecell{N-Gram, XGB\\Walk}
\\
\hline
\hline
Delaney&RMSE
&0.866 &0.746 &0.769 &\underline{\textbf{0.731}} \\
\hline
Malaria&RMSE
&1.036 &1.027 &1.022 &\underline{\textbf{1.019}} \\
\hline
CEP&RMSE
&1.506 &\underline{\textbf{1.350}} &1.399 &1.366 \\
\hline
QM7&MAE
&73.745 &57.361 &57.747 &\underline{\textbf{53.919}} \\
\hline
E1-CC2&MAE
&0.011 &0.009 &0.008 &\underline{\textbf{0.007}} \\
E2-CC2&MAE
&0.011 &0.009 &0.009 &\underline{\textbf{0.008}} \\
f1-CC2&MAE
&0.017 &0.016 &0.015 &\underline{\textbf{0.015}} \\
f2-CC2&MAE
&0.036 &0.034 &0.033 &\underline{\textbf{0.031}} \\
E1-PBE0&MAE
&0.011 &0.009 &0.008 &\underline{\textbf{0.007}} \\
E2-PBE0&MAE
&0.010 &0.009 &0.008 &\underline{\textbf{0.008}} \\
f1-PBE0&MAE
&0.015 &0.014 &0.013 &\underline{\textbf{0.013}} \\
f2-PBE0&MAE
&0.028 &0.027 &0.025 &\underline{\textbf{0.024}} \\
E1-CAM&MAE
&0.010 &0.008 &0.007 &\underline{\textbf{0.007}} \\
E2-CAM&MAE
&0.010 &0.008 &0.008 &\underline{\textbf{0.007}} \\
f1-CAM&MAE
&0.017 &0.015 &0.014 &\underline{\textbf{0.014}} \\
f2-CAM&MAE
&0.031 &0.030 &0.028 &\underline{\textbf{0.026}} \\
average& &0.017 &0.016 &0.015 &0.014 \\
\hline
mu&MAE
&0.629 &0.588 &0.562 &\underline{\textbf{0.535}} \\
alpha&MAE
&0.868 &0.759 &0.722 &\underline{\textbf{0.612}} \\
homo&MAE
&0.006 &0.006 &0.005 &\underline{\textbf{0.005}} \\
lumo&MAE
&0.007 &0.006 &0.006 &\underline{\textbf{0.005}} \\
gap&MAE
&0.009 &0.008 &0.007 &\underline{\textbf{0.007}} \\
r2&MAE
&88.431 &67.876 &72.846 &\underline{\textbf{59.137}} \\
zpve&MAE
&0.001 &0.001 &0.001 &\underline{\textbf{0.000}} \\
cv&MAE
&0.613 &0.498 &0.434 &\underline{\textbf{0.334}} \\
u0&MAE
&1.382 &0.592 &0.429 &\underline{\textbf{0.427}} \\
u298&MAE
&1.384 &0.594 &0.429 &\underline{\textbf{0.428}} \\
h298&MAE
&1.380 &0.591 &0.428 &\underline{\textbf{0.428}} \\
g298&MAE
&1.382 &0.593 &0.428 &\underline{\textbf{0.428}} \\
average& &8.035 &6.001 &6.357 &5.152 \\
\hline
    \end{tabular}
\end{table}
\newpage
\begin{table}[H]
    \small
    \caption{
        Comparison of N-Gram Path-based graph and N-Gram Walk-based graph on 32 classification tasks.
        All experiments are done on a 5-fold cross-validation, and the mean evaluation of 5 runs is reported here.
        N-Gram Walk-based graph with RF and XGB trained on top of it can excel on 5 and 15 tasks out of 32 respectively.
        Besides, the average performance trained with RF and XGB is better when using N-Gram walk-based graph, except for XGB on HIV. Overall, N-Gram Walk is better while N-Gram Path achieves comparable performance. 
    }
    \label{tab:walk_path_classification_results}
    \centering
    \begin{tabular}{|C{2.5cm}|c|c|c|c|c|}
\hline
Task&Eval Metric&\makecell{N-Gram, RF\\Path}&\makecell{N-Gram, XGB\\Path}&\makecell{N-Gram, RF\\Walk}&\makecell{N-Gram, XGB\\Walk}
\\
\hline
\hline
NR-AR&ROC-AUC
&\underline{\textbf{0.797}} &0.788 &0.797 &0.791 \\
NR-AR-LBD&ROC-AUC
&0.860 &0.857 &\underline{\textbf{0.871}} &0.864 \\
NR-AhR&ROC-AUC
&0.890 &0.896 &0.894 &\underline{\textbf{0.902}} \\
NR-Aromatase&ROC-AUC
&0.856 &0.863 &0.858 &\underline{\textbf{0.869}} \\
NR-ER&ROC-AUC
&0.742 &0.750 &0.747 &\underline{\textbf{0.753}} \\
NR-ER-LBD&ROC-AUC
&0.823 &\underline{\textbf{0.840}} &0.827 &0.838 \\
NR-PPAR-gamma&ROC-AUC
&0.837 &0.832 &\underline{\textbf{0.856}} &0.851 \\
SR-ARE&ROC-AUC
&0.824 &0.834 &0.826 &\underline{\textbf{0.835}} \\
SR-ATAD5&ROC-AUC
&0.858 &0.848 &0.857 &\underline{\textbf{0.860}} \\
SR-HSE&ROC-AUC
&0.790 &0.795 &0.798 &\underline{\textbf{0.812}} \\
SR-MMP&ROC-AUC
&0.904 &0.912 &0.911 &\underline{\textbf{0.918}} \\
SR-p53&ROC-AUC
&0.847 &0.850 &0.859 &\underline{\textbf{0.868}} \\
average& &0.833 &0.834 &0.841 &0.842 \\
\hline
CT\_TOX&ROC-AUC
&0.838 &0.858 &0.857 &\underline{\textbf{0.873}} \\
FDA\_APPROVED&ROC-AUC
&0.816 &0.854 &0.825 &\underline{\textbf{0.874}} \\
average& &0.810 &0.855 &0.837 &0.870 \\
\hline
MUV-466&PR-AUC
&0.056 &0.077 &0.058 &\underline{\textbf{0.086}} \\
MUV-548&PR-AUC
&0.088 &\underline{\textbf{0.100}} &0.073 &0.094 \\
MUV-600&PR-AUC
&0.008 &\underline{\textbf{0.014}} &0.007 &0.009 \\
MUV-644&PR-AUC
&0.061 &\underline{\textbf{0.093}} &0.046 &0.064 \\
MUV-652&PR-AUC
&0.096 &\underline{\textbf{0.151}} &0.085 &0.118 \\
MUV-689&PR-AUC
&0.027 &0.025 &0.026 &\underline{\textbf{0.046}} \\
MUV-692&PR-AUC
&0.004 &0.003 &\underline{\textbf{0.005}} &0.005 \\
MUV-712&PR-AUC
&0.088 &0.085 &0.134 &\underline{\textbf{0.151}} \\
MUV-713&PR-AUC
&\underline{\textbf{0.052}} &0.015 &0.026 &0.026 \\
MUV-733&PR-AUC
&0.017 &0.015 &0.021 &\underline{\textbf{0.047}} \\
MUV-737&PR-AUC
&0.038 &0.035 &\underline{\textbf{0.084}} &0.080 \\
MUV-810&PR-AUC
&0.017 &\underline{\textbf{0.054}} &0.013 &0.022 \\
MUV-832&PR-AUC
&0.176 &\underline{\textbf{0.297}} &0.229 &0.280 \\
MUV-846&PR-AUC
&0.245 &0.223 &\underline{\textbf{0.250}} &0.220 \\
MUV-852&PR-AUC
&0.188 &0.189 &0.214 &\underline{\textbf{0.238}} \\
MUV-858&PR-AUC
&0.007 &\underline{\textbf{0.016}} &0.014 &0.015 \\
MUV-859&PR-AUC
&\underline{\textbf{0.012}} &0.005 &0.007 &0.006 \\
average& &0.079 &0.087 &0.088 &0.099 \\
\hline
HIV&ROC-AUC
&0.826 &\underline{\textbf{0.833}} &0.828 &0.830 \\
\hline
    \end{tabular}
\end{table}

\newpage

\section{Additional Experiments on Datasets with 3D Information} \label{s:3d_results}


Since 3D information of the atoms in the molecules is important for making predictions \cite{gilmer2017neural}, we also performed experiments comparing out method to two recent models designed to exploit 3D information: Deep Tensor Neural Networks (DTNN) \cite{schutt2017quantum} and Message-Passing Neural Networks (MPNN) \cite{gilmer2017neural}.
We evaluated them on the two datasets QM8 and QM9 that have 3D information.

The detailed results are in \Cref{tab:performance_regression_3d} and the summary is in \Cref{tab:performance_3D}.
The computational time can be referred to \Cref{tab:representation_construction_time_3d}.
The results show that our method, though not using 3D information, can get comparable performance. 

\begin{table}[H]
    \scriptsize
    \centering
    \caption{
        Here we include the performance on 2 regression datasets with 9 models.
        All experiments are done on a 5-fold cross-validation, and the mean evaluation of 5 runs is reported here.
        The top-3 models are \textbf{bolded}, and the best model is \underline{\textbf{underlined}}.
    }
    \label{tab:performance_regression_3d}
    \begin{tabular}{|c|c|c|c|c|c|c|c|c|c|c|}
\hline
Task&Eval Metric&\makecell{WL\\SVM}&\makecell{Morgan\\RF}&\makecell{Morgan\\XGB}&GCNN&Weave&DTNN&MPNN&\makecell{N-Gram\\RF}&\makecell{N-Gram\\XGB}
\\
\hline
\hline
E1-CC2&MAE
&0.032 &0.008 &0.008 &\textbf{0.006} &0.007 &\textbf{0.006} &\underline{\textbf{0.006}} &0.008 &0.007 \\
E2-CC2&MAE
&0.023 &0.010 &0.010 &0.008 &\textbf{0.007} &\textbf{0.007} &\underline{\textbf{0.007}} &0.009 &0.008 \\
f1-CC2&MAE
&0.072 &\underline{\textbf{0.014}} &\textbf{0.015} &\textbf{0.014} &0.018 &0.021 &0.019 &0.015 &0.015 \\
f2-CC2&MAE
&0.081 &\textbf{0.032} &0.033 &\textbf{0.031} &0.036 &0.042 &0.038 &0.033 &\underline{\textbf{0.031}} \\
E1-PBE0&MAE
&0.034 &0.008 &0.008 &0.006 &\textbf{0.006} &\textbf{0.006} &\underline{\textbf{0.006}} &0.008 &0.007 \\
E2-PBE0&MAE
&0.029 &0.010 &0.010 &\textbf{0.007} &0.008 &\textbf{0.007} &\underline{\textbf{0.006}} &0.008 &0.008 \\
f1-PBE0&MAE
&0.068 &\textbf{0.012} &0.013 &\underline{\textbf{0.012}} &0.014 &0.018 &0.016 &0.013 &\textbf{0.013} \\
f2-PBE0&MAE
&0.078 &0.026 &0.027 &\textbf{0.024} &0.027 &0.035 &0.030 &\textbf{0.025} &\underline{\textbf{0.024}} \\
E1-CAM&MAE
&0.033 &0.007 &0.007 &\underline{\textbf{0.006}} &0.006 &\textbf{0.006} &\textbf{0.006} &0.007 &0.007 \\
E2-CAM&MAE
&0.025 &0.009 &0.009 &\textbf{0.006} &\textbf{0.006} &0.007 &\underline{\textbf{0.006}} &0.008 &0.007 \\
f1-CAM&MAE
&0.073 &\textbf{0.013} &0.014 &\underline{\textbf{0.013}} &0.016 &0.019 &0.018 &0.014 &\textbf{0.014} \\
f2-CAM&MAE
&0.080 &0.028 &0.028 &\underline{\textbf{0.026}} &0.031 &0.037 &0.032 &\textbf{0.028} &\textbf{0.026} \\
average& &0.052 &0.015 &0.015 &0.013 &0.015 &0.018 &0.016 &0.015 &0.014 \\
\hline
mu&MAE
&-- &0.548 &0.533 &\textbf{0.482} &0.624 &\underline{\textbf{0.238}} &\textbf{0.308} &0.562 &0.535 \\
alpha&MAE
&-- &3.787 &2.672 &0.685 &1.034 &\underline{\textbf{0.445}} &\textbf{0.621} &0.722 &\textbf{0.612} \\
homo&MAE
&-- &0.006 &0.006 &\textbf{0.004} &0.005 &\underline{\textbf{0.003}} &\textbf{0.004} &0.005 &0.005 \\
lumo&MAE
&-- &0.007 &0.006 &\textbf{0.004} &0.005 &\underline{\textbf{0.004}} &\textbf{0.004} &0.006 &0.005 \\
gap&MAE
&-- &0.008 &0.008 &\textbf{0.006} &0.008 &\underline{\textbf{0.005}} &\textbf{0.006} &0.007 &0.007 \\
r2&MAE
&-- &94.815 &82.516 &64.775 &\textbf{42.095} &\textbf{10.405} &\underline{\textbf{10.198}} &72.846 &59.137 \\
zpve&MAE
&-- &0.009 &0.007 &0.001 &0.002 &\underline{\textbf{0.000}} &0.001 &\textbf{0.001} &\textbf{0.000} \\
cv&MAE
&-- &1.505 &1.166 &0.524 &0.374 &\underline{\textbf{0.132}} &\textbf{0.241} &0.434 &\textbf{0.334} \\
u0&MAE
&-- &16.410 &12.736 &2.460 &1.465 &1.142 &\textbf{0.866} &\textbf{0.429} &\underline{\textbf{0.427}} \\
u298&MAE
&-- &16.410 &12.757 &2.671 &1.560 &1.838 &\textbf{0.991} &\textbf{0.429} &\underline{\textbf{0.428}} \\
h298&MAE
&-- &16.411 &12.752 &2.542 &1.414 &\textbf{0.737} &1.146 &\textbf{0.428} &\underline{\textbf{0.428}} \\
g298&MAE
&-- &16.414 &12.750 &2.466 &2.359 &\textbf{0.853} &1.166 &\textbf{0.428} &\underline{\textbf{0.428}} \\
average& &nan &13.823 &11.476 &6.474 &4.187 &1.328 &1.187 &6.357 &5.152 \\
\hline
    \end{tabular}
\end{table}

\begin{table}[htb!]
\centering
\scriptsize
    \caption{
        Representation construction time in seconds.
        One task from each dataset as an example.
        Average over 5 folds, and including both the training set and test set.
    }
    \label{tab:representation_construction_time_3d}
    \begin{tabular}{|c|c||c|c||c|c|c|c|c||c|c|}
\hline
Task&Dataset&\makecell{WL\\CPU}&\makecell{Morgan FPs\\CPU}&\makecell{GCNN\\GPU}&\makecell{Weave\\GPU}&\makecell{DTNN\\GPU}&\makecell{MPNN\\GPU}&\makecell{GIN\\GPU}&\makecell{Vertex\\ Emb\\GPU}&\makecell{Graph\\Emb\\GPU}
\\
\hline
\hline
Delaney&Delaney&2.46&0.25&39.70&65.82&--&124.89&--&49.63&2.90 \\
Malaria&Malaria&128.81&5.28&377.24&536.99&--&--&--&1152.80&19.58 \\
CEP&CEP&1113.35&17.69&607.23&849.37&--&--&--&2695.57&37.40 \\
qm7&qm7&60.24&0.98&103.12&76.48&--&--&--&173.50&10.60 \\
E1-CC2&qm8&584.98&3.60&382.72&262.16&928.61&2431.28&--&966.49&33.43 \\
mu&qm9&--&19.58&9051.37&1504.77&6275.30&10770.84&--&8279.03&169.72 \\
NR-AR&tox21&70.35&2.03&130.15&142.59&--&--&608.57&525.24&10.81 \\
CT-TOX&clintox&4.92&0.63&62.61&95.50&--&--&135.68&191.93&3.83 \\
MUV-466&muv&276.42&6.31&401.02&690.15&--&--&1327.26&1221.25&25.50 \\
hiv&hiv&2284.74&17.16&1142.77&2138.10&--&--&3641.52&3975.76&139.85 \\
\hline
    \end{tabular}
\end{table}

\newpage

\section{Exploring the Effects of $r$ and $T$} \label{s:effect_hyper}

\subsection{On 12 Classification Tasks (Tox12)}
We run N-gram graph on 12 classification tasks from "Toxicology in the 21st Century" \cite{tox21}.
We tested the effects of vertex embedding dimension $r$ and N-gram parameter $T$ on the prediction performance measured by ROC-AUC. The results are shown in \Cref{fig:all_tox21}.

As observed from \Cref{fig:all_tox21}, for the 12 tasks from Tox21, there generally exists a raise as $T$ gets higher. This makes sense since it covers more information as we are looking more steps ahead.
Besides, the ROC-AUC values on the test set are not increasing as $r$ increases.
Two possible reasons for this:
(1) Data is insufficient. As shown in \Cref{tab:tox21_distribution,tab:clintox_distribution,tab:muv_distribution,tab:hiv_distribution}, all Tox21 tasks have less than 10,000 molecules.
(2) ROC-AUC reveals the ranking of predictions, while other evaluation metrics, like RMSE shown in \Cref{fig:all_rmse}, are likely to measure the predictions in a finer-grained way.



\subsection{On 3 Regression Tasks (Delaney, Malaria, CEP)}

We run N-gram graph on 3 regression tasks, Delaney, Malaria, and CEP.
We tested the effects of vertex embedding dimension $r$ and N-gram parameter $T$.

\begin{figure}[htb!]
    \centering
    \includegraphics[width=1.\linewidth]{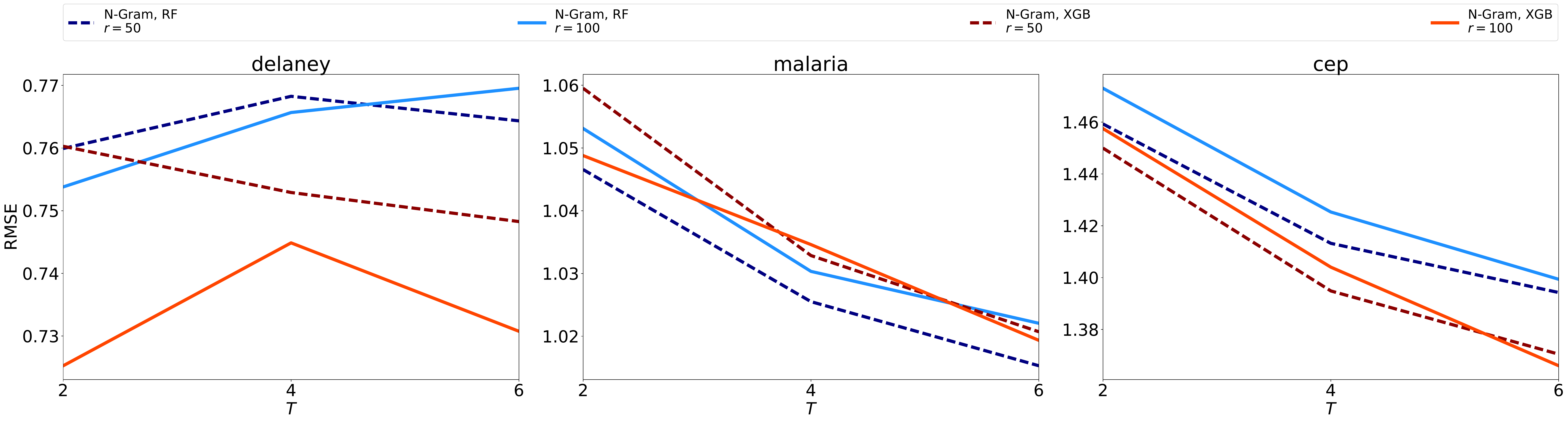}
    \caption{Effects of vertex embedding dimension $r$ and N-gram dimension $T$ on tasks Delaney, Malaria and CEP: how the RMSE on validation set changes as different $r$ and $T$.}
    \label{fig:all_rmse}
\end{figure}

Similarly to \Cref{fig:all_tox21}, increasing $T$ can help reduce the loss, while different vertex embedding dimension, \ie $r$, presents comparatively unstable performance.
Performance on the three regression tasks in \Cref{fig:all_rmse} fluctuates a lot as $r$ and $T$ increases.
One conjecture is that such high variance is caused by the data insufficiency.
However, we can still conclude that for each machine learning algorithm, $r=100$ and $T=6$ are reasonable to choose.

\section{Exploring the Effects of Atom Features} \label{sec:feature_effects}

To further prove that different atom features are not biasing the graph neural networks, we compare the different atom attribute schemes.
In N-Gram graph, we are using \Cref{tab:vertex_attribute_matrix} (called new attribute scheme), while in the benchmark paper \cite{wu2018moleculenet}, it has more atom symbols, and may not include attributes like ``is acceptor'' or ``is donor'' (called original attribute scheme). We did a statistical test to measure the difference from two atom attribute schemes as in \Cref{tab:tukey_diff_embeddings}.

\begin{table}[H]
    \centering
    \caption{
    For each message-passing graph method, we compare the performance on 12 Tox21 tasks.
    The null hypothesis here is that means are the same, so rejection=False means we should accept the null hypothesis.
    Thus, this table shows that two attribute schemes contain very similar information with respect to the performance.
    }
    \label{tab:tukey_diff_embeddings}
    \begin{tabular}{|c|c|c|c|}
        \hline
         Group 1 & Group 2 & mean diff & reject  \\
         \hline
         GCNN new attribute scheme & GCNN original attribute scheme & -0.0012 & False\\
         Weave new attribute scheme & Weave original attribute scheme & 0.0008 & False\\
         \hline
    \end{tabular}
\end{table}

\end{document}